\documentclass{article}

\PassOptionsToPackage{numbers, compress}{natbib}

\usepackage[final]{neurips_data_2023}

\usepackage[utf8]{inputenc} 
\usepackage[T1]{fontenc}    
\usepackage{hyperref}       
\usepackage{url}            
\usepackage{booktabs}       
\usepackage{amsfonts}       
\usepackage{nicefrac}       
\usepackage{microtype}      
\usepackage{xcolor}         

\usepackage{amsmath}
\usepackage{amssymb}
\usepackage{mathtools}
\usepackage{amsthm}
\usepackage{defs_ahn}
\usepackage{multirow}
\usepackage{hyperref}
\usepackage{url}
\usepackage{graphicx}
\usepackage{booktabs}
\usepackage[export]{adjustbox}
\usepackage{color,soul}
\usepackage{xargs}
\usepackage{caption}
\usepackage{subcaption}
\usepackage{lipsum}
\usepackage{listings}
\usepackage{placeins}
\usepackage{xspace}
\usepackage{pifont}
\usepackage{colortbl}
\usepackage{wrapfig}
\usepackage{algorithm}
\usepackage{algorithmic}
\usepackage{dirtree}

\usepackage{enumitem}
\setlist{leftmargin=15pt,labelindent=15pt}

\usepackage{rotating}
\usepackage{arydshln}
\usepackage{array}
\usepackage{setspace}

\newcommand{\cmark}{\textcolor{teal}{\ding{51}}}
\newcommand{\xmark}{\textcolor{purple}{\ding{55}}}
\newcommand{\tmark}{\textcolor{orange}{\ding{115}}}

\definecolor{mygreen}{RGB}{0,128,0}
\definecolor{mylimegreen}{RGB}{50,205,50}
\definecolor{myyellowgreen}{RGB}{154,205,50}
\definecolor{myyellow}{RGB}{255,255,0}

\lstdefinelanguage{json}{
    basicstyle=\normalfont\ttfamily,
    commentstyle=\color{gray},
    stringstyle=\color{blue},
    literate=
     *{:}{{{\color{purple}{:}}}}{1}
      {,}{{{\color{purple}{,}}}}{1}
      {\{}{{{\color{brown}{\{}}}}{1}
      {\}}{{{\color{brown}{\}}}}}{1}
      {[}{{{\color{brown}{[}}}}{1}
      {]}{{{\color{brown}{]}}}}{1},
}

\newcolumntype{L}[1]{>{\raggedright\let\newline\\\arraybackslash\hspace{0pt}}m{#1}}
\newcolumntype{C}[1]{>{\centering\let\newline\\\arraybackslash\hspace{0pt}}m{#1}}
\newcolumntype{R}[1]{>{\raggedleft\let\newline\\\arraybackslash\hspace{0pt}}m{#1}}

\title{Imagine the Unseen World: A Benchmark for Systematic Generalization in Visual World Models}

\author{%
   Yeongbin Kim\thanks{Equal contribution.} \\ KAIST
   \And
   Gautam Singh$^{*}$ \\ Rutgers University 
   \And
   Junyeong Park \\ KAIST 
   \And
   Caglar Gulcehre\thanks{This work was partly done while C.G. was at Google DeepMind. C.G. is currently affiliated with EPFL.} \\ EPFL \& Google DeepMind 
   \And
   Sungjin Ahn\thanks{Correspondence to \texttt{sungjin.ahn@kaist.ac.kr}.}  \\ KAIST 
}

\begin{document}

\maketitle

\begin{abstract}
Systematic compositionality, or the ability to adapt to novel situations by creating a mental model of the world using reusable pieces of knowledge, remains a significant challenge in machine learning. While there has been considerable progress in the language domain, efforts towards systematic visual imagination, or envisioning the dynamical implications of a visual observation, are in their infancy. We introduce the \textit{Systematic Visual Imagination Benchmark} (SVIB), the first benchmark designed to address this problem head-on. SVIB offers a novel framework for a minimal world modeling problem, where models are evaluated based on their ability to generate one-step image-to-image transformations under a latent world dynamics. The framework provides benefits such as the possibility to jointly optimize for systematic perception and imagination, a range of difficulty levels, and the ability to control the fraction of possible factor combinations used during training. We provide a comprehensive evaluation of various baseline models on SVIB, offering insight into the current state-of-the-art in systematic visual imagination. We hope that this benchmark will help advance visual systematic compositionality. \url{https://systematic-visual-imagination.github.io}\footnote{This is the official project page. It provides links to the benchmark and the code.}
\end{abstract}

\section{Introduction}
Constructing a mental model of the world, known as a world model, in a composable way is a crucial aspect of human intelligence~\citep{worldmodels,whatiscognitivemap,mattar22}. This ability enables humans to adapt to novel situations and problems in a zero-shot manner by envisioning the possible future~\citep{pearson2019human,bakermans2023constructing}. Studies in neuroscience and cognitive science suggest that the key to this ability is the process of acquiring abstract, conceptual, and reusable pieces of knowledge from past experiences and applying them in new configurations to comprehend a novel situation~\citep{fodor88,bernardi2020geometry,olafsdottir2018role,bakermans2023constructing}. For instance, a person who understands the implications of a scene involving a ``big dog'' and a ``small cat'', e.g., how they may interact and what the subsequent scene may look like, can also reasonably understand the implications of an unfamiliar scenario involving a ``big cat'' and a ``small dog''. While this capability, termed systematic compositionality, is fundamental to human intelligence, how neural networks can acquire such an ability remains one of the grand challenges in machine learning~\citep{scan_lake,bahdanau2018systematic,onbinding}. 

The notion of systematic compositionality originates from the fields of philosophy of language and linguistics~\citep{compositionalitystanford,fodor88}. Language inherently provides a compositional representation using token structures, such as words or characters, which simplifies addressing compositional systematicity. As a result, the AI community has also made efforts to address this problem predominantly in the language domain~\citep{devlin2017robustfill,scan_lake,kim2020cogs,shi2022compositional,ling2017program,amini2019mathqa,keysers2019measuring,finegan2018improving,shaw2020compositional, wu2023recogs}. One notable milestone driving progress in this field recently is the development of the SCAN benchmark~\citep{scan_lake} posing a sequence-to-sequence translation task. The authors have demonstrated that RNNs fail catastrophically at test time when presented with a \textit{compositionally novel} input, i.e., an unknown composition of known concepts. 

However, the problem is even more elusive when it comes to imagining the dynamical implications of a visual observation, a problem referred to in this work as the \textit{compositional or systematic visual imagination}. One reason for this is that, unlike language, images do not naturally provide such a token-based compositional representation, making the problem significantly more challenging. To tackle this problem, one must not only learn how to utilize compositional representations for systematic composition but also obtain such representations from unstructured, complex, high-dimensional pixels. Another reason is the lack of an appropriate benchmark to directly address it. Although several prior works have explored related problems, none have tackled it head-on.

Specifically, the prior research related to systematic visual imagination can be grouped into three categories. The first group~\citep{ruis2020benchmark, wu2021reascan} tackles this issue by providing language alongside images, e.g., Visual Question Answering (VQA)~\citep{clevr,bahdanau2018systematic,bogin2021covr, bahdanau2019closure,grunde2021agqa}, thus pursuing an indirect task that bypasses the challenge of learning compositional representation by leveraging the inherent token-based compositional nature of language. The second group~\citep{schmidhuber1996semilinear, bengio2013representation, higgins2017scan, zhao2018bias, montero2021role, xu2022compositional} also focuses on an indirect task, i.e., obtaining disentangled representations.~The underlying hypothesis is that a disentangled representation such as those from variational autoencoders~\citep{vae, higgins2017beta} or object-centric representations~\citep{slotattention,monet,space,scalor, wu2022slotformer,slate,steve,sysbinder,dinosour} will naturally lead to a solution to the systematic generalization problem as well. However, it restricts exploring potential solutions not relying on disentanglement \cite{coat}, and recent studies suggest that disentanglement may not necessarily lead to systematic generalization~\citep{montero2021role, xu2022compositional}. The third group focuses on visual reasoning~\citep{chollet2019measure, webb2020learning, pekar2020generating, zhang2021abstract, assouel2022object} while ignoring the problem of visual perceptual systematic generalization, essential for zero-shot problem-solving. For instance, the two closest visual reasoning benchmarks to ours, ARC~\citep{chollet2019measure} and Sort-of-ARC~\citep{assouel2022object}, exhibit these limitations. ARC lacks suitability for evaluating systematic generalization due to its non-procedural generation and lack of access to underlying component factors and rules. Sort-of-ARC, on the other hand, emphasizes inferring the underlying rule composition from a few-shot support set while ignoring the zero-shot systematic visual imagination ability---a cornerstone of our proposed benchmark.

In this paper, we introduce a new benchmark called the \textit{Systematic Visual Imagination Benchmark} (SVIB), the first benchmark to rigorously address the compositional visual imagination problem. The SVIB poses the task as a minimal world modeling problem: one-step image-to-image generation. The objective is to generate the subsequent scene image from the current one. The underlying world dynamics operate as a relational function of object-centric factors. The benchmark presents a subset of possible combinations of the visual primitives during training, and during testing, exclusively presents compositionally novel input images to assess a model's systematic imagination ability.

SVIB offers numerous benefits for studying the compositional visual imagination problem. Our proposed task framework provides a means to jointly optimize for systematic perception and imagination, instead of optimizing an indirect objective such as disentanglement or object segmentation quality. 
Unlike many works in disentanglement literature focusing on a single-object scene~\citep{higgins2017scan, montero2021role, xu2022compositional, van2019disentangled}, SVIB incorporates multi-object scenes where each object is constructed via composition rules in terms of well-defined factors like color, shape, and size. SVIB also provides various difficulty levels, accommodating visual complexity from simplistic 2D images to realistically textured 3D scene images, and varying complexity of the world dynamics. Additionally, SVIB provides control over the fraction of possible factor combinations used in training, thus serving as a systematic generalization difficulty meter. The simplicity of one-step generation allows us to concentrate on the systematic imagination ability, sidestepping the computational demands of long-range generation abilities that video generation tasks often entail. 

Our paper makes two main contributions:  First, we propose the first benchmark to foster research on the systematic or compositional visual imagination problem, providing several unique benefits as outlined above. Second, we conduct comprehensive empirical evaluations of various baseline models on SVIB, shedding light on the current state-of-the-art capabilities in this field.

\section{SVIB: Systematic Visual Imagination Benchmark}
\label{sec:method}

\begin{figure*}[t]
    \centering
    \includegraphics[width=\textwidth]{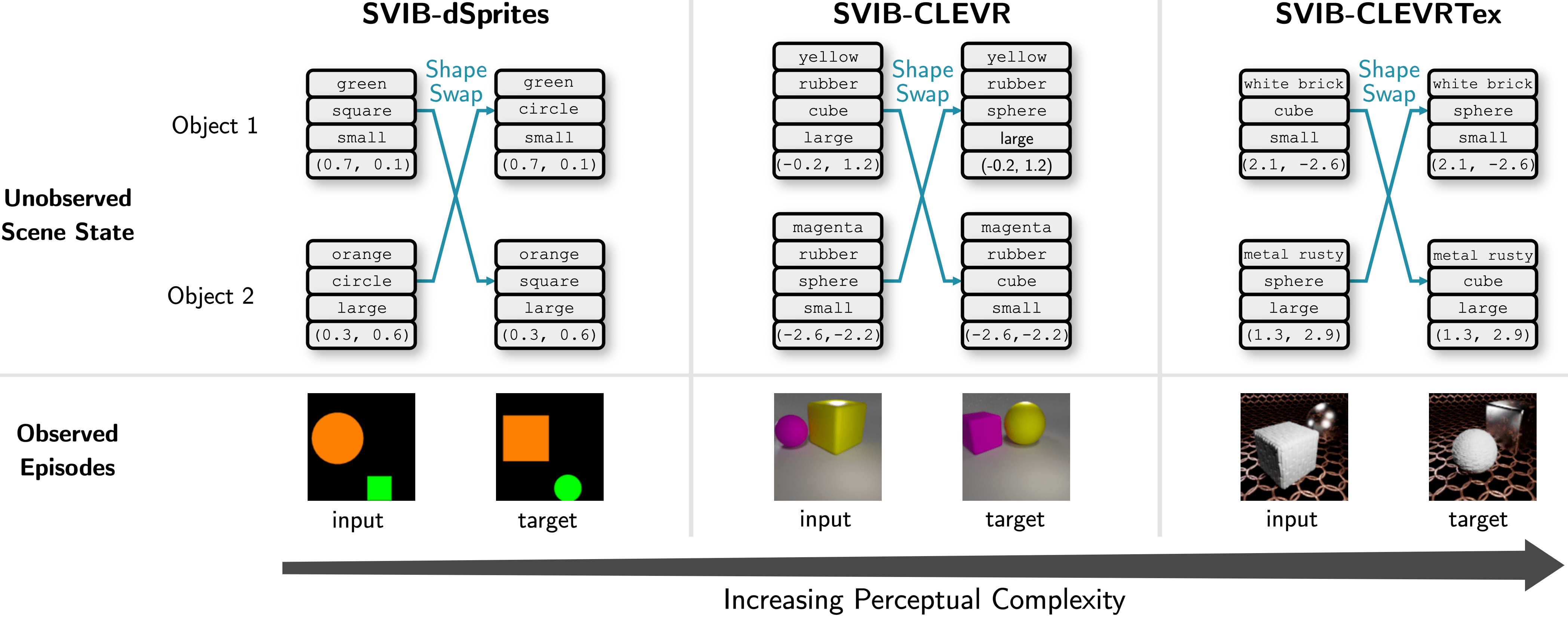}
    \caption{\textbf{Compositional Visual World in SVIB}: Our benchmark provides 2-frame videos of an underlying compositional visual world. The scenes contain two objects, each composed using intra-object factors such as color, shape, size, material, etc. In this figure, we show episodes following the \textit{Shape-Swap} rule i.e., the input state is transformed to the target state by swapping the shapes of the two objects. 
    We illustrate episodes from three tasks having different visual complexity levels.}
    \label{fig:overview_of_benchmark_main}
\end{figure*}

In this section, we describe the \textit{Systematic Visual Imagination Benchmark} or \textit{SVIB} that we propose. The benchmark is designed to train models to perform visual imagination and measure their systematic generalization ability. Our benchmark provides a catalog of 12 \textit{visual imagination tasks}\footnote{See our project page at \url{https://systematic-visual-imagination.github.io}.}. 
Each \textit{task} contains 64K episodes for training a model and 8K systematically out-of-distribution episodes for testing the model. Each episode is a two-frame video as illustrated in Fig.~\ref{fig:overview_of_benchmark_main}. We denote the two frames as the input image $\bx$ and the target image $\by$. Using the training episodes, a model must learn to map the input image to its corresponding target image. The model is then evaluated on its ability to accurately predict target images from the input images in testing episodes, which present systematically out-of-distribution scenes.

\subsection{Compositional Visual World}
For each task, the episodes are generated from an underlying compositional visual world defined by \textit{1)} a \textit{visual vocabulary}---a library of visual primitives that act as building blocks for constructing the scenes, and \textit{2)} a \textit{rule} that determines the mapping from the input scene to the target scene. 

\textbf{Scene Compositionality and Visual Vocabulary.} In SVIB, each scene is composed of two objects. The objects are composed of intra-object \textit{factors} such as color, shape, size, \textit{etc.} These factors take their values from a collection of visual primitives called a \textit{visual vocabulary}. For example, if an object can be described by its color, shape, and size then we can define a visual vocabulary containing colors $\{$\texttt{green}, \texttt{blue}, \texttt{magenta}, \texttt{orange}$\}$, shapes $\{$\texttt{circle}, \texttt{triangle}, \texttt{square}, \texttt{star}$\}$, and sizes $\{$\texttt{tiny}, \texttt{small}, \texttt{medium}, \texttt{large}$\}$. Under a given vocabulary, we refer to any composition of factors that completely describes an object's appearance as a \textit{combination}. Given our example vocabulary, two possible combinations are: \texttt{blue-circle-tiny} and \texttt{orange-triangle-large}. By placing the objects defined by these combinations inside a scene, we can construct an example scene containing a \texttt{blue-circle-tiny} at position $(0.1, 0.4)$ and a \texttt{orange-triangle-large} at position $(0.7, 0.1)$.

\textbf{Rule.} A \textit{rule} is a function that transforms the input scene state to the target scene state. In SVIB, the rule is fixed for all episodes within a task. The rule modifies specific factor values of both objects in the scene. The rule is symmetric, i.e., the same rule is executed on both objects. An example of a rule in SVIB is the \textit{Shape-Swap} rule that swaps the shapes of the two objects to transform the input scene state to the target scene state. In SVIB, we identify two axes of rule complexity and our tasks span these two complexity axes. The first axis is \textit{the number of modified factors per object}. We call a rule that modifies only a single factor per object as a \textit{single} rule. If a rule modifies multiple factors per object, we call it a \textit{multiple} rule. The second axis is \textit{the number of parents} i.e., the number of factors that determine the modified factor value. If only one parent is involved per factor, we call the rule an \textit{atomic} rule else we call it a \textit{non-atomic} rule. With these two axes, we define 4 rule categories with increasing complexity: \textit{single atomic}, \textit{multiple atomic}, \textit{single non-atomic}, and \textit{multiple non-atomic}. For example, the aforementioned Shape-Swap rule can be categorized as \textit{single atomic} because only one factor (i.e., shape) is modified per object whose new value depends on only one parent, i.e., the shape of the other object.

\begin{figure*}[t]
    \centering
    \includegraphics[width=1.0\textwidth]{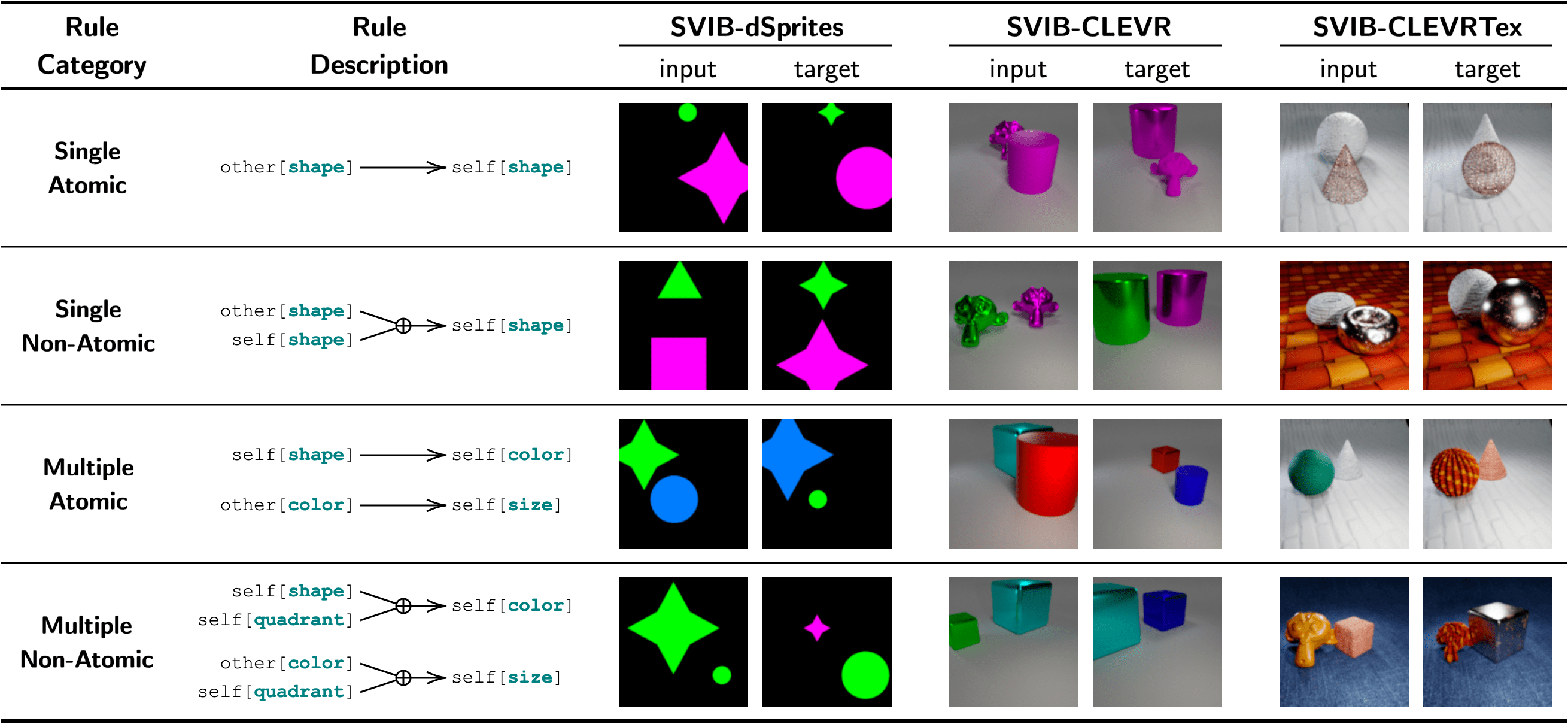}
     \caption{\textbf{SVIB Tasks and the Transformation Rules.} In this illustration, we show sample episodes of all 12 tasks in our benchmark. For each instance, we show the input and target images. We describe the underlying rule that governs the transformation of the input image to the target image. In the rule description diagrams, each factor value is considered to be an integer index in its corresponding vocabulary. A direct arrow indicates a simple assignment operation and a `$\oplus$' symbol indicates a summation operation. All assignments are performed modulo the vocabulary size of the target factor. As the rules are applied symmetrically to both objects, we describe the rules with respect to the \texttt{self} object while \texttt{other} denotes the other object.}
    \label{fig:rule-description-main}
\end{figure*}

\subsection{Systematic Training and Testing Splits}
In this section, we describe how we construct the training and the testing episodes of a task. As a benchmark for studying systematic generalization, our task episodes satisfy the following three conditions: \textit{1)} each primitive in the visual vocabulary is shown individually in the training episodes, \textit{2)} a subset of combinations is reserved for testing and \textit{3)} the fraction of combinations exposed during training can be controlled, offering a control knob to adjust the difficulty of generalization. To satisfy these, we proceed as follows.

\textbf{Core Combinations and Testing Combinations.} We first construct a set of \textit{core combinations}---the smallest set of combinations that contains each primitive in the visual vocabulary at least once. For instance, for our previous example of a visual vocabulary, we can define the set of core combinations as: $\{$\texttt{green-circle-tiny}, \texttt{blue-square-small}, \texttt{magenta-square-medium}, \texttt{orange-star-large}$\}$. After defining the core combinations, we reserve 20\% of the remaining combinations as \textit{testing combinations}. For more details, see Appendix \ref{ax:core}.

\begin{figure*}[t]
    \centering
    \includegraphics[width=\textwidth]{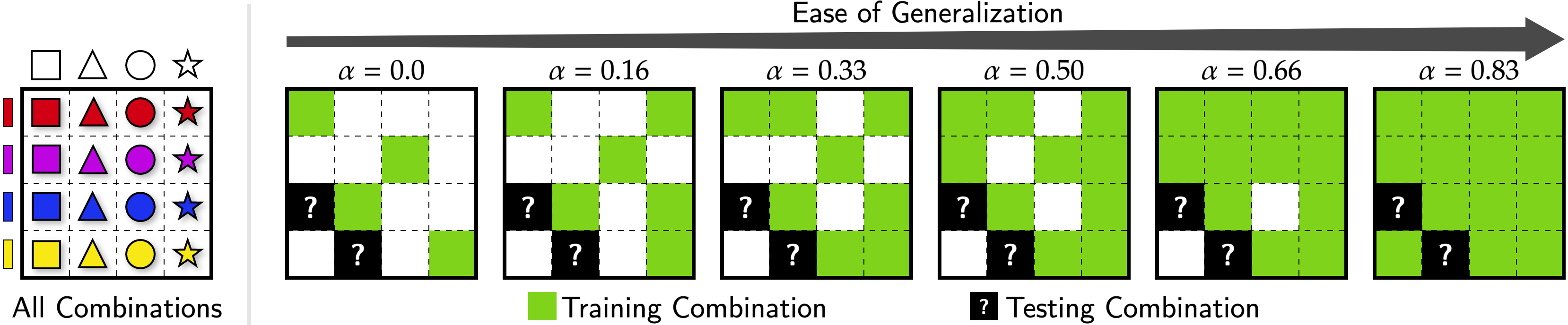}
     \caption{\textbf{The $
     \alpha$-Rating of Training Splits in SVIB.} \textit{Left:} In this illustration, we showcase a library of 4 shape primitives and 4 color primitives that result in a total of 16 possible combinations or objects. \textit{Right:} In training episodes, we present each shape and color primitive individually, however, we do not show all possible combinations. To control the ease of solving the generalization task, we define an $\alpha$-rating of the training split, the fraction of the combinations presented during training. A higher $\alpha$-rating corresponds to more combinations shown in the training split. In testing episodes, we present only the held-out combinations. 
    }
    \label{fig:alpha}
\end{figure*}

\textbf{The $\alpha$-Rating and Training Combinations.} Each training split in our benchmark is associated with an \textit{$\alpha$-rating}. To create a training split having a specific $\alpha$-rating, we randomly select an $\alpha$ fraction of the combinations that remain after setting aside the core and the testing combinations. These selected combinations are then added to the core combinations to obtain the set of \textit{training combinations}. The $\alpha$-rating acts as a control knob over how difficult it is to generalize for a model trained on this split. A higher $\alpha$-rating corresponds to exposing more combinations in the training split, thereby providing an easier generalization setting. Similarly, a low $\alpha$-rating corresponds to a more difficult generalization setting. In Figure \ref{fig:alpha}, we provide an illustration. 

\textbf{Generating the Episodes.} To generate a training episode, we randomly select two combinations from the training set and randomly position the objects defined by these combinations to construct an input scene. Next, the task rule is applied to this input scene to generate the target scene. The input and the target scenes are rendered to generate the two frames of the episode. By sampling episodes repeatedly, we obtain the complete training split. The testing split is generated analogously from the set of testing combinations. 

\subsection{Contents of the Benchmark}
\label{sec:reference_tasks}

\textbf{Tasks.} SVIB comprises 12 tasks, divided into three subsets with four tasks each, categorized by the visual complexity of their respective visual worlds: \textit{1)} {SVIB-dSprites} is based on visually simple 2D scenes; \textit{2)} {SVIB-CLEVR} is based on visually simple 3D scenes; and \textit{3)} {SVIB-CLEVRTex} is based on 3D scenes with complex textures. This gradation in visual complexity is illustrated in Figure \ref{fig:overview_of_benchmark_main}. Within each of these three subsets, the 4 tasks correspond to 4 rules with increasing rule complexity: \textit{Single Atomic} (S-A), \textit{Single Non-Atomic} (S-NA), \textit{Multiple Atomic} (M-A), and \textit{Multiple Non-Atomic} (M-NA). These 4 rules are illustrated in Figure \ref{fig:rule-description-main}. For their formal definitions, see Appendix \ref{ax:reference-rule-defs}. For each rule, we provide 3 training splits with distinct $\alpha$-ratings denoted as \textit{Easy} split ($\alpha=0.6$), \textit{Medium} split ($\alpha=0.4$) and \textit{Hard} split ($\alpha=0.2$) and a common testing split. Each training split contains 64K episodes and the testing split contains 8K episodes. For each episode, we provide the input image, the target image, the ground-truth scene descriptions, and the ground-truth object masks. In Figure \ref{fig:dir_tree}, we show the complete directory structure of our benchmark.

\textbf{Omni-Composition Datasets.} SVIB additionally provides an \textit{omni-composition dataset} for each of the 3 visual worlds i.e., {SVIB-dSprites}, {SVIB-CLEVR}, and {SVIB-CLEVRTex}. An omni-composition dataset is a dataset containing unpaired images that expose all possible combinations of primitives under the visual vocabulary of its respective visual world. The use of omni-composition datasets for solving the benchmark tasks is optional.

\section{Metrics}

\textbf{MSE.} A natural metric for measuring the systematic generalization performance of a model is the mean squared error (MSE) between predicted and target images on the testing set episodes (denoted as $\smash{\text{MSE}^\text{OOD}}$). 
We can also compute how much worse the OOD MSE is relative to the in-distribution MSE via the \textit{systematic generalization gap}: $\smash{\log \text{MSE}^\text{OOD} - \log \text{MSE}^\text{ID}}$, where $\smash{\text{MSE}^\text{ID}}$ denotes the in-distribution MSE.

\textbf{LPIPS.} A key limitation of MSE is that it can overly focus on low-level errors, e.g., small deviations that do not significantly impact human perception. Therefore we recommend reporting LPIPS~\citep{zhang2018unreasonable}, an error metric shown to better correspond to human judgments. LPIPS works by taking two images as input, extracting their features via a pre-trained image classification model, and computing a weighted distance between the extracted features. In our experiments, we use the official implementation\footnote{\url{https://github.com/richzhang/PerceptualSimilarity}}.

\section{Baselines}
\label{ref:baselines}

In this section, we describe the baselines that we evaluate in our experiments. These were selected considering two main implications of our benchmark: 
evaluating systematic generalization in vision models and world models.
We evaluate two categories of baselines: \textit{1)} Image-to-Image Models, and \textit{2)} State-Space Models. Additionally, we evaluate an \emph{Oracle} model to obtain the best-case performance and to set a milestone of success on the benchmark tasks. 

\subsection{Image-to-Image Models}

In this category, we consider end-to-end neural network models that take the input image $\bx$ and try to predict the target image $\by$. Within the model, the input image is mapped to an intermediate hidden representation $\bee$ which is used to generate the prediction $\hat{\by}$. This can be described as follows: 
\begin{align*}
\bee = \text{Encoder}_\ta(\bx) && \Longrightarrow && \hat{\by} = \text{Decoder}_\gamma(\bee)
\end{align*}
We test two variants that capture the two extremes of how expressive the intermediate representation $\bee$ is. The first variant employs a CNN as the encoder to map the input image to single-vector representation $\bee$,  thus creating a bottleneck of low expressiveness. 
We denote this baseline as \textit{Image-to-Image CNN} or I2I-CNN. The second variant employs a Vision Transformer or ViT as the encoder and produces a representation that is a collection of multiple vectors---as many as the patches in the image \citep{vit}. We denote this baseline as \textit{Image-to-Image ViT} or I2I-ViT. 

\textbf{Oracle.} To set a milestone of success on our benchmark, we construct a baseline where instead of employing an image encoder to obtain the representation $\bee$, we directly stack the vector representations of the ground truth scene factors and provide it to the decoder. For more details, see Appendix \ref{ax:oracle}.

\subsection{State-Space Models}
State-Space Models (or SSMs) are a category of models that first encode the input image $\bx$ to a latent representation $\bz_\bx$ and then apply a latent-level dynamics model to predict the target scene latent.
\begin{align*}
   \bz_\bx = \text{Encoder}_\phi(\bx) && \Longrightarrow && \hat{\bz}_\by = \text{Dynamics}_\ta(\bz_\bx) && \Longrightarrow && \hat{\by} = \text{Probe}_\gamma(\hat{\bz}_\by),
\end{align*}
where $\bz_\bx$, $\hat{\bz}_\by \in \cZ$. 
We test two variants that employ two different kinds of latent structures. The first variant adopts Variational Auto-Encoding or VAE for obtaining latent scene representations---a popular representation method for world modeling \citep{vae, planet, dreamer}. The encoded latent is a single vector where factors of scene variation are expected to be disentangled and captured by individual dimensions of the vector. We denote this variant as \textit{SSM-VAE}. The second variant adopts Slot Attention Video or SAVi---a popular object-centric representation method \citep{slotattention, kipf2021conditional, wu2022slotformer}. SAVi represents each frame by a set of $N$ object vectors, also known as slots. Furthermore, the slot order is preserved across frames, facilitating the training of a slot-level dynamics model. We denote this variant as \textit{SSM-Slot}. Finally, since it is not possible to compute the MSE or LPIPS performance from the predicted latent, we train a probe $\text{Probe}_\gamma(\cdot)$ that takes the predicted latent $\hat{\bz}_\by$ and decodes it to generate the target image $\hat{\by}$.
To ensure a fair performance comparison, we keep the probe's architecture to be the same as that of the decoder used in the Image-to-Image baselines and the Oracle. For more details, see Appendix \ref{sec:transformer_decoder}.

\section{Related Work}

\textbf{Systematic Generalization in Language.} In the language domain, a substantial number of studies, including SCAN~\citep{scan_lake}, have examined systematic generalization through sequence-to-sequence tasks that translate different sequence forms~\citep{devlin2017robustfill,kim2020cogs,shi2022compositional,ling2017program,amini2019mathqa,keysers2019measuring,finegan2018improving,shaw2020compositional, wu2023recogs}. Some other benchmarks like gSCAN~\citep{ruis2020benchmark} and ReaSCAN~\citep{wu2021reascan} utilize additional inputs such as images and videos alongside language~\citep{qiu2021systematic,sleeper2022grounded,ma2022crepe}. In \textit{Visual Question Answering} or VQA, studies test systematicity by providing images with unseen object combinations~\citep{clevr, bahdanau2018systematic} or presenting questions in unseen formats~\citep{bogin2021covr, bahdanau2019closure,grunde2021agqa}. However, all of these either test systematic generalization in language or utilize language inputs when testing visual compositionality.

\textbf{Representation Learning.} 
Another group of works questions specific representation learning methodologies and whether they support systematic generalization.
One line of representation methods seeks disentangled single-vector representations \citep{higgins2017beta, Kim2018DisentanglingBF, Kumar2018VariationalIO}.
However, studies that investigate it either do not quantify (e.g., via $\alpha$-rating) the degree to which OOD inputs systematically differ from the training \cite{van2019disentangled, xu2022compositional} or tackle only visually simplistic and single-object scenes \cite{van2019disentangled, xu2022compositional, Trauble2020OnDR, montero2021role, montero2022lost}, unlike ours. Furthermore, \citep{higgins2017darla}, relying on RL, lacks the simplicity of our one-step prediction framework. 
Higgins \emph{et al.}~\citep{higgins2017scan} test the importance of disentanglement in a symbol-to-image task, however, the learning is partly supervised by symbol labels.
Zhao \emph{et al.}~\citep{zhao2018bias} study generative models (e.g., VAE) and whether the learned distributions capture OOD regions of the observation space, however, they lack a focus on perception unlike ours. 
Another line of representation methods seeks object-centric representations \cite{tagger, air, space, iodine, genesis, slotattention, slate, dinosour}. Although these are considered promising to support OOD generalization \cite{onbinding, sysbinder}, very few studies investigate this potential. Such studies tend to lack well-defined and systematic factor recombination for OOD testing \cite{dittadi2021generalization, Dittadi2022TheRO, jaesikocrl, Mambelli2022CompositionalMR, cobra} or rely on RL \cite{Dittadi2022TheRO, jaesikocrl, Mambelli2022CompositionalMR, cobra}, lacking the simplicity of one-step prediction.

\textbf{Visual Reasoning.} In the visual domain, several studies test reasoning abilities by inferring the underlying rule from a support set and then finding images from a candidate set that conform to that rule~\citep{barrett2018measuring, zhang2019raven, hu2021stratified, benny2021scale, hill2019learning, teney2020v, ichien2021visual}. Alternatively, there are odd-one-out tasks that find the rule-violating image from an input image set without requiring a separate support set~\citep{zerroug2022benchmark, mandziuk2019deepiq} and tasks that categorize input images into several groups~\citep{fleuret2011comparing, gulccehre2016knowledge, vedantam2021curi, nie2020bongard, jiang2022bongard}. However, their learning setup is discriminative and not generative, unlike ours. 
Some reasoning tasks require learning to generate~\citep{chollet2019measure, webb2020learning, pekar2020generating, zhang2021abstract}, however, these tasks fall outside the scope of visual systematic generalization. Assouel \emph{et al.}~\citep{assouel2022object} improve upon the ARC benchmark \citep{chollet2019measure} by properly defining elementary primitives and procedurally generating the dataset to make it suitable to study systematic generalization. However, the primitives and their combinations are concerned with scene dynamics and not the visual observation itself, unlike ours. Furthermore, both benchmarks \citep{chollet2019measure, assouel2022object} use coarse and simplistic observations, whereas ours provides visually complex high-resolution images.

\section{Experiments}

\begin{figure*}[t]
    \centering
    \includegraphics[width=1.0\textwidth]{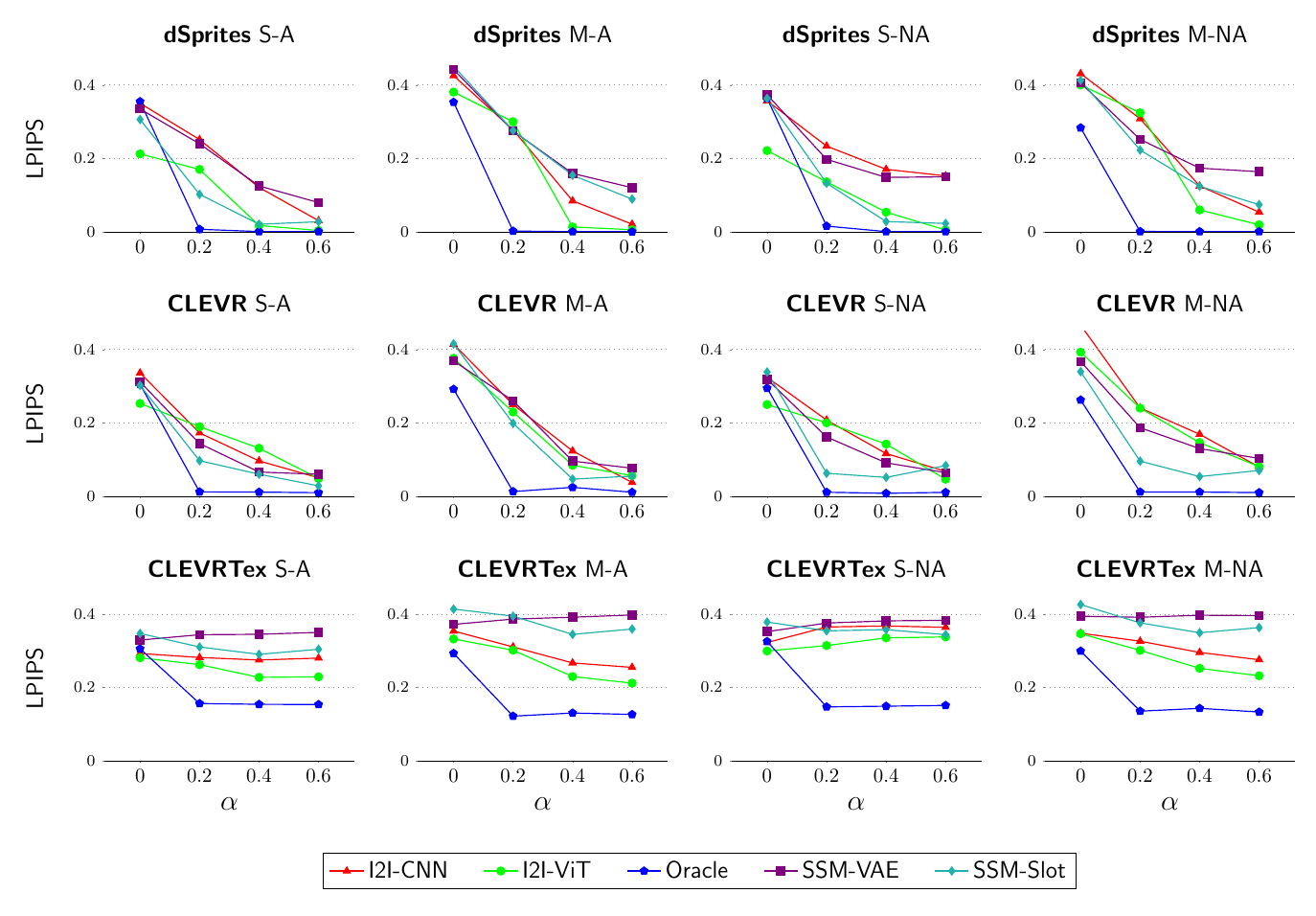}
    \caption{\textbf{Systematic Generalization in Visual Imagination.} We plot LPIPS with respect to $\alpha$ for all benchmark tasks and baselines on the systematically out-of-distribution (OOD) test set. Lower is better. For SVIB-dSprites, SVIB-CLEVR, and SVIB-CLEVRTex, we evaluate the tasks: single atomic (S-A), single non-atomic (S-NA), multiple atomic (M-A), and multiple non-atomic (M-NA).
    }
    \label{fig:all-reference-lpips}
\end{figure*}

\begin{figure*}[t]
    \centering
    \includegraphics[width=0.95\textwidth]{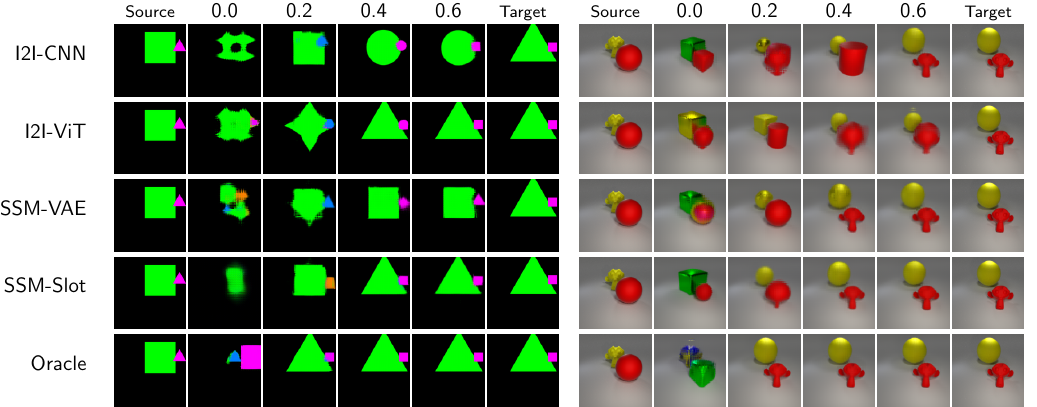}
    \caption{\textbf{Qualitative Performance of the baselines on SVIB-dSprites and SVIB-CLEVR.} We illustrate the predictions made by various baselines on the Shape-Swap task, where the task is to swap the shapes of the two objects in the input image. We show predictions on a test episode for various baselines for varying $\alpha$ of the training split.}
    \label{fig:qualitative}
\end{figure*}

We test the five baselines as outlined in Section \ref{ref:baselines}: Image-to-Image CNN, Image-to-Image ViT, SSM-VAE, SSM-Slot, and Oracle on the 12 benchmark tasks outlined in Section~\ref{sec:reference_tasks}.
In Figures~\ref{fig:all-reference-lpips},~\ref{fig:all-reference-mse}, and~\ref{fig:all-reference-sgg}, we report the LPIPS, the MSE, and the systematic generalization gaps, respectively.
Since LPIPS better reflects the human judgment of qualitative performance, we base our conclusions in this section on LPIPS. In our analysis, we consider Oracle's performance as a yardstick for task success. 
In Fig.~\ref{fig:qualitative} and \ref{fig:qualitative_clevrtex}, we show samples of the model predictions.


\subsection{Effect of \texorpdfstring{$\alpha$}{Alpha} on Systematic Generalization}
\textbf{Performance at $\alpha=0.0$.}  When $\alpha=0.0$, the training split only includes the core combinations. In all tasks, we note that all baselines fail completely for $\alpha=0.0$. This is supported by the qualitative results in Fig.~\ref{fig:qualitative}, where most predicted images are garbled blobs. This can be expected since at $\alpha=0.0$, the factors are highly correlated in the training split, making it difficult for the learner to infer the true causal parent in the rule. For instance, in the Shape-Swap task in SVIB-dSprites: the learner sees only the core combinations:  $\{$\texttt{green-circle-tiny}, \texttt{blue-square-small}, \texttt{magenta-square-medium}, \texttt{orange-star-large}$\}$ where every shape value is always shown with a specific color value and a specific size value, making it impossible (even for the Oracle) to determine whether the true causal parent is color, shape, or size. 

\textbf{Performance with Increasing $\alpha$.} As $\alpha$ is increased from $0.0$, the difficulty of systematic generalization becomes less severe and we see a consistent downward slope in the error plots across baselines and tasks. This is also supported by the qualitative results in Fig.~\ref{fig:qualitative}, where, with increasing $\alpha$, a greater number of baselines are able to improve their predictions. Since training splits with larger $\alpha$ expose more combinations, there is more opportunity for the baselines to overcome learning spurious correlations and discover the compositional primitives and the causal rule to apply. 

\textbf{Performance at $\alpha=0.6$.} Our easiest training split corresponds to $\alpha=0.6$ which exposes the largest fraction of combinations during training. Yet, on most tasks, most baselines are not able to reach the performance of Oracle---although it does happen in some select cases. From this, we can say that our benchmark is largely unsolved with significant room for progress. 

\textbf{Performance of Oracle.} We note that Oracle is the best performer as it is the only baseline that can solve the tasks as early as at $\alpha=0.2$. In comparison to Oracle, other baselines show a much flatter downward slope. This can be explained by considering that SVIB requires systematic perception i.e., learning to perceive the input in terms of useful tokens---explicitly or implicitly---while also generalizing systematically to OOD inputs. Oracle is designed with access to the ground-truth tokens directly and it does not need to learn systematic perception via training. On the contrary, other baselines carry the additional burden of learning systematic perception along with rule learning---contributing to their worse performance. 

\textbf{Performance on SVIB-CLEVRTex.} The performance of baselines on SVIB-CLEVRTex requires a special mention as we see that the curves with respect to $\alpha$ are almost flat and the slope is not as large as that in the other simpler environments i.e., SVIB-dSprites and SVIB-CLEVR. That is, the baselines struggle to systematically generalize in SVIB-CLEVRTex. A notable exception is Oracle which can solve the tasks starting at $\alpha=0.2$. What distinguishes Oracle from other baselines is its systematic perception ability. As such, we can attribute the degraded performance of other baselines in SVIB-CLEVRTex to their poor systematic perception.
This shows that higher visual complexity can make systematic perception more difficult in comparison to visually simple scenes of SVIB-dSprites and SVIB-CLEVR. This observation is also supported by the predictions visualized in Fig.~\ref{fig:qualitative_clevrtex}. It is noteworthy that our proposed benchmark provides a unique opportunity to study systematic imagination in visually complex scenes as the existing image-to-image benchmarks provide visually toy-like scenes \citep{chollet2019measure, assouel2022object}.

\renewcommand{\arraystretch}{1.2}
\begin{table}[t]
\centering
\caption{{\textbf{LPIPS Performance on SVIB.} This table summarizes the LPIPS performance results of various models evaluated across all 12 tasks of SVIB. The values are computed by averaging the results from four tasks for each difficulty level.}}
\resizebox{\textwidth}{!}{
\begin{tabular}{L{1.8cm}@{\hskip 0.0in}C{1.5cm}@{\hskip 0.0in}@{\hskip 0.0in}C{1.5cm}@{\hskip 0.0in}@{\hskip 0.0in}C{1.5cm}@{\hskip 0.0in}@{\hskip 0.0in}C{1.5cm}@{\hskip 0.0in}@{\hskip 0.0in}C{1.5cm}@{\hskip 0.0in}@{\hskip 0.0in}C{1.5cm}@{\hskip 0.0in}@{\hskip 0.0in}C{1.5cm}@{\hskip 0.0in}@{\hskip 0.0in}C{1.5cm}@{\hskip 0.0in}@{\hskip 0.0in}C{1.5cm}@{\hskip 0.0in}}
\noalign{\smallskip}\noalign{\smallskip}
\toprule
            & \multicolumn{3}{c}{\textbf{SVIB-dSprites}} & \multicolumn{3}{c}{\textbf{SVIB-CLEVR}} & \multicolumn{3}{c}{\textbf{SVIB-CLEVRTex}}\\
\cmidrule(lr){2-4}\cmidrule(lr){5-7}\cmidrule(lr){8-10}
Models      & Easy              & Medium                & Hard              & Easy              & Medium            & Hard      & Easy      & Medium    & Hard      \\
\midrule

I2I-CNN     &0.0643             &0.1251                 &0.2678             &0.0593             &0.1265             &0.2181         &0.2938             &0.3014             &0.3210        \\
I2I-ViT     &\textbf{0.0084}    &\textbf{0.0359}        &0.2326             &\textbf{0.0583}    &0.1263             &0.2153     &\textbf{0.2529}    &\textbf{0.2614}    &\textbf{0.2950} \\
SSM-VAE     &0.1284             &0.1515                 &0.2410             &0.0763             &0.0961    &0.1884         &0.3819             &0.3789             &0.3745         \\
SSM-Slot    &0.0534             &0.0817                 &\textbf{0.1834}    &0.0596             &\textbf{0.0533}             &\textbf{0.1139}    &0.3429             &0.3357             &0.3590         \\
\cmidrule(lr){1-10} 
Oracle      &0.0006             &0.0006                 &0.0064             &0.0107             &0.0141             &0.0122         &0.1414             &0.1444             &0.1404          \\

\bottomrule

\end{tabular}}
\label{table:all-reference-lpips}
\end{table}
\renewcommand{\arraystretch}{1}

\subsection{Comparison of Baselines}
We now compare the baselines in more detail. We draw our conclusions based on tasks with ratings of $\alpha = 0.2$ and $0.4$ since the tasks are too difficult for $\alpha=0.0$ and, in select cases, too easy for $\alpha=0.6$.

\textbf{Image-to-Image Models.} In Image-to-Image models, we compare two encoders: CNN and ViT.  In SVIB-dSprites, we note that ViT is generally superior to CNN. 
This is also evidenced in the qualitative results of SVIB-dSprites in Fig.~\ref{fig:qualitative} where ViT can correctly predict for $\alpha \geq 0.4$ while CNN continues to predict the wrong object shapes on all $\alpha$ values. 
It is somewhat surprising that ViT, which lacks any explicit design to encourage systematic generalization, can generalize at all.
However, in SVIB-CLEVR and SVIB-CLEVRTex, the performances of CNN and ViT are comparable. This suggests that although ViT is more capable than CNN on SVIB-dSprites, the greater visual complexity of SVIB-CLEVR and SVIB-CLEVRTex can make ViT struggle similarly to CNN.

\textbf{State-Space Models.} Comparing SSM-VAE and SSM-Slot, we find that the SSM-Slot generally outperforms the SSM-VAE. This is further evidenced by the qualitative results of SVIB-dSprites in Fig.~\ref{fig:qualitative} where SSM-VAE predicts an incorrect shape even at $\alpha=0.6$ while SSM-Slot is able to predict correctly for $\alpha \geq 0.4$. However, both SSMs still remain significantly worse than the Oracle.

\textbf{State-Space Modeling versus Image-to-Image Modeling.} 
We compare our best-performing image-to-image model with our best-performing state-space model i.e., I2I-ViT versus SSM-Slot. 
In SVIB-dSprites, both models perform comparably. We believe this is due to the visual simplicity of SVIB-dSprites which allows ViT to learn systematic perception implicitly without any explicit inductive biases such as those imposed on the SSMs.
In SVIB-CLEVR, SSM-Slot seems to generalize significantly better than I2I-ViT. 
This is also evidenced in the qualitative results in Fig.~\ref{fig:qualitative}, where SSM-Slot can correctly generate the red monkey for $\alpha \geq 0.4$ while I2I-ViT fails for all $\alpha$ values. 
We hypothesize that since SSMs enforce an inductive bias on the latent, this can eventually help in learning systematic perception.
In SVIB-CLEVRTex, both baselines fail completely. In fact, SSM-Slot fails worse than I2I-ViT. We think this is due to a familiar problem of the conventional Slot Attention-based models that they cannot extract objects properly in visually complex scenes \citep{karazija2021clevrtex, steve}.

\subsection{Summary of Experiment Results}
We summarize the main experimental findings  and pointers for making progress on our benchmark:
\enums{
\itemsep0em 
\item All baselines fail completely at $\alpha=0.0$. Although most baselines demonstrate improved performance with increasing $\alpha$, only a select few can eventually solve some of the tasks and reach the performance of the Oracle. This highlights the challenging nature of our tasks and indicates a substantial opportunity for improvement.
\item Systematic perception is central to solving our tasks. However, visual complexity can make perception and, as a result, task-solving much more difficult. Our benchmark provides a unique opportunity to study systematic generalization along the axis of visual complexity.
\item Excluding the Oracle, in SVIB-dSprites and SVIB-CLEVR, ViT outperforms all other baselines on the \textit{Easy} and \textit{Medium} training splits but loses to the object-centric baseline SSM-Slot on the \textit{Hard} training split. In SVIB-CLEVRTex, ViT outperforms all other baselines but significantly falls short of reaching the successful performance of the Oracle.
\item Models based on single-vector representations (e.g., I2I-CNN and SSM-VAE) tend to perform worse than their multi-vector counterparts (e.g., I2I-ViT and SSM-Slot).
}

\section{Conclusion}

We introduced the \textit{Systematic Visual Imagination Benchmark} (SVIB)---a novel image-to-image generation benchmark that focuses on visual compositionality and systematic generalization. SVIB is procedurally generated and contains multi-object scenes, with each object compositionally constructed from well-defined visual primitives. The underlying image-to-image mapping rule is designed to require extracting the underlying visual factors.
Our experiments showed that our benchmark is yet to be solved and highlights an important limitation of the current models. We also showed that systematic perception and visual complexity are important aspects of this problem.
We hope that our benchmark will enable the development of more capable world modeling and perception approaches.

\textbf{Limitations and Future Extensions.} Our benchmark makes several simplifying choices to support faster model development and ease of analysis. However, this leaves several limitations and avenues for extending the benchmark.
First, from a world modeling perspective, a richer benchmark may be constructed by introducing action-conditioning, stochasticity, longer episodes, occlusions, 3D viewpoints, \textit{etc.}
Second, our dynamics are fixed within a task. Our benchmark can be extended to have distinct dynamics in each episode constructed compositionally using dynamical primitives. 
Third, our benchmark can be extended by introducing greater realism e.g., greater visual complexity, more objects, more primitives, \textit{etc.} 
Fourth, our dynamics involve symmetric rules and a future extension may introduce non-symmetric rules. Another avenue is to increase emphasis on relational abstractions.
Lastly, future works may also consider investigating whether large pre-trained models (e.g.,~\cite{mae}) can provide necessary priors for systematic generalization.

\section*{Acknowledgements and Disclosure of Funding}
This work is supported by Brain Pool Plus Program (No. 2021H1D3A2A03103645) through the National Research Foundation of Korea (NRF) funded by the Ministry of Science and ICT.

\bibliography{refs}
\bibliographystyle{plain}

\clearpage

\appendix

\section{Datasheet for Datasets}
\label{ax:datasheet}
\subsection{Motivation}

\subsubsection*{Question 1: For what purpose was the dataset created?}

The dataset was created as a test-bed to evaluate the systematic generalization ability of visual imagination models, with an emphasis on compositionality at the level of intra-object factors (e.g., color, shape, size, etc.) and visual complexity.

\subsubsection*{Question 2: Who created the dataset (e.g., which team or research group) and on behalf of which entity (e.g., company, institution, organization)?}

The dataset was created by the authors who are affiliated with Machine Learning and Mind Lab (MLML) situated in the School of Computing at Korea Advanced Institute of Science and Technology (KAIST) and Rutgers University.

\subsubsection*{Question 3: Who funded the creation of the dataset?} 

This work is supported by Brain Pool Plus Program (No. 2021H1D3A2A03103645) through the National Research Foundation of Korea (NRF) funded by the Ministry of Science and ICT.

\subsection{Composition}

\subsubsection*{Question 1: What do the instances that comprise the dataset represent (e.g., documents, photos, people, countries)?} The dataset comprises several tasks. Within each task, we provide 1) several training splits of different difficulties of systematic generalization, and 2) a test split. Within each split, we provide several pairs of images, with the first image in each pair serving as the input and the second image serving as the prediction target. These images are purely synthetic and procedurally generated using the Blender\footnote{\url{https://www.blender.org}} and the Spriteworld API\footnote{\url{https://github.com/deepmind/spriteworld}}. The images showcase multi-object scenes where the objects have simple shapes (e.g., sphere, cube), simple colors (e.g., red, blue), standard materials (e.g., brick, rubber, metal), and multiple sizes (e.g., small, medium, large). No real data (e.g., about people or countries) was collected or used in our data creation process. Along with the images, we also provide scene meta-data files detailing the configurations of the objects, lighting, camera, and background of the scene. 
We also provide the ground truth object masks.

\subsubsection*{Question 2: How many instances are there in total of each type?}

\begin{figure}[t]
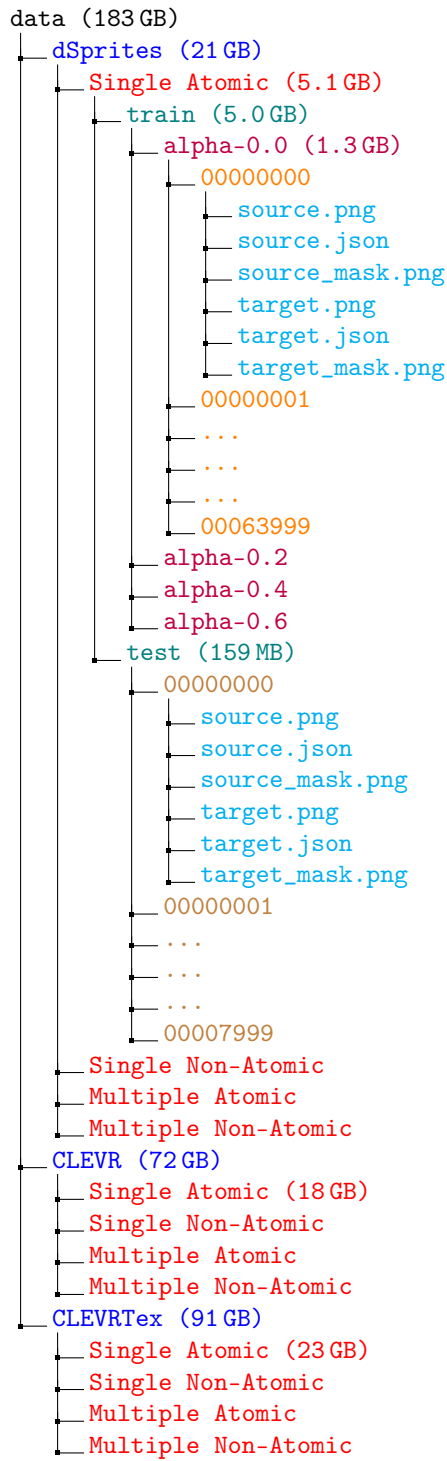

\centering
\begin{minipage}{0.5\textwidth} 
\renewcommand{\DTbaselineskip}{12pt}
\dirtree{%
.1 \textbf{data (183\,GB)}.
.2 \textcolor{blue}{\textbf{dSprites} (21\,GB)}.
.3 \textcolor{red}{Single Atomic (5.1\,GB)}.
.4 \textcolor{teal}{train (5.0\,GB)}.
.5 \textcolor{purple}{alpha-0.0 (1.3\,GB)}.
.6 \textcolor{orange}{00000000}.
.7 \textcolor{cyan}{source.png}.
.7 \textcolor{cyan}{source.json}.
.7 \textcolor{cyan}{source\_mask.png}.
.7 \textcolor{cyan}{target.png}.
.7 \textcolor{cyan}{target.json}.
.7 \textcolor{cyan}{target\_mask.png}.
.6 \textcolor{orange}{00000001}.
.6 \textcolor{orange}{...}.
.6 \textcolor{orange}{...}.
.6 \textcolor{orange}{...}.
.6 \textcolor{orange}{00063999}.
.5 \textcolor{purple}{alpha-0.2}.
.5 \textcolor{purple}{alpha-0.4}.
.5 \textcolor{purple}{alpha-0.6}.
.4 \textcolor{teal}{test (159\,MB)}.
.5 \textcolor{brown}{00000000}.
.6 \textcolor{cyan}{source.png}.
.6 \textcolor{cyan}{source.json}.
.6 \textcolor{cyan}{source\_mask.png}.
.6 \textcolor{cyan}{target.png}.
.6 \textcolor{cyan}{target.json}.
.6 \textcolor{cyan}{target\_mask.png}.
.5 \textcolor{brown}{00000001}.
.5 \textcolor{brown}{...}.
.5 \textcolor{brown}{...}.
.5 \textcolor{brown}{...}.
.5 \textcolor{brown}{00007999}.
.3 \textcolor{red}{Single Non-Atomic}.
.3 \textcolor{red}{Multiple Atomic}.
.3 \textcolor{red}{Multiple Non-Atomic}.
.2 \textcolor{blue}{\textbf{CLEVR} (72\,GB)}.
.3 \textcolor{red}{Single Atomic (18\,GB)}.
.3 \textcolor{red}{Single Non-Atomic}.
.3 \textcolor{red}{Multiple Atomic}.
.3 \textcolor{red}{Multiple Non-Atomic}.
.2 \textcolor{blue}{\textbf{CLEVRTex (91\,GB)}}.
.3 \textcolor{red}{Single Atomic (23\,GB)}.
.3 \textcolor{red}{Single Non-Atomic}.
.3 \textcolor{red}{Multiple Atomic}.
.3 \textcolor{red}{Multiple Non-Atomic}.
}
\end{minipage}
\caption{Directory structure of the dataset. We show a detailed view of the SVIB-dSprites directory. The structure of other environments i.e., SVIB-CLEVR and SVIB-CLEVRTex are identical. We also show the directory sizes on the disk in parentheses next to the directory names.}
\label{fig:dir_tree}
\end{figure}

In our dataset, there are a total of 12 tasks, 4 tasks per environment (SVIB-dSprites, SVIB-CLEVR, and SVIB-CLEVRTex). Within each task, we provide 4 training splits corresponding to $\alpha$ values $0.0$, $0.2$, $0.4$, and $0.6$. The training splits $0.2$, $0.4$, and $0.6$ are denoted as Hard, Medium, and Easy, respectively. Within each training split, we provide 64000 input and target images. Furthermore, within a task, we provide a test split containing 8000 out-of-distribution input and target images. We illustrate the complete directory structure of our dataset in Figure \ref{fig:dir_tree}.

\subsubsection*{Question 3: Does the dataset contain all possible instances or is it a sample (not necessarily random) of instances from a larger set?}

For each intra-object factor (e.g., color or shape), we define a library of primitives from which the factor takes values. These libraries of primitives are finite meaning that these libraries are not an exhaustive list of all possible values that a factor can theoretically take. Nevertheless, our libraries are designed to capture all standard factor values as also done in the previous works \citep{clevr, karazija2021clevrtex}.

\subsubsection*{Question 4: What data does each instance consist of?}

Each instance consists of 4 files: \texttt{source.png}, \texttt{source.json}, \texttt{target.png}, \texttt{target.json}. The \texttt{source.png} and \texttt{target.png} are PNG files containing the $128\times 128$-sized source and target images, respectively. The \texttt{source.json} and \texttt{target.json} are JSON files detailing the scene meta-data of the source and the target scenes, respectively. See Figure~\ref{fig:json_structure}
 for an illustration of the JSON contents.
\begin{figure}[t]
\centering
\scriptsize
\begin{minipage}{0.49\textwidth}
\begin{lstlisting}[language=json]
{
  "image_filename": "00000000",
  "objects": [
    {
      "shape": "triangle", 
      "size": 0.325, 
      "rotation": 0.0,
      "2d_coords": [0.629, 0.365],
      "color": [255, 0, 255]
    },
    {
      "shape": "circle", 
      "size": 0.425, 
      "rotation": 0.0,
      "2d_coords": [0.684, 0.797],
      "color": [0, 127, 255]
    }
  ]
}
\end{lstlisting}
\end{minipage}
\begin{minipage}{0.49\textwidth}
\begin{lstlisting}[language=json]
{
  "image_filename": "000000110120338217",
  "objects": [
    {
      "color": [1.0, 1.0, 0.0, 1.0],
      "material": "MyMetal",
      "3d_coords": [2.924, -2.007, 2.0],
      "pixel_coords": [77, 62, 7.241],
      "size": 2.0,
      "shape": "Sphere",
      "rotation": 0.0
    },
    {
      "color": [0.0, 0.0, 1.0, 1.0],
      "material": "Rubber",
      "3d_coords": [-2.835, 1.059, 2.0],
      "pixel_coords": [50, 30, 12.723],
      "size": 2.0,
      "shape": "SmoothCylinder",
      "rotation": 0.0
    }
  ],
  "Camera": [6.990, -6.999, 5.379],
  "Lamp_Back": [-1.111, 2.506, 6.118],
  "Lamp_Key": [6.451, -3.099, 4.898],
  "Lamp_Fill": [-3.825, -3.888, 2.036]
}
\end{lstlisting}
\end{minipage}
\caption{Examples of JSON files describing the ground truth scene structure of an SVIB-dSprites image (left) and an SVIB-CLEVR image (right). Within each of the images, there are two objects. Each object has factors such as shape, size, rotation, 2D coordinates, 3D coordinates, and RGBA color values. In SVIB-CLEVR, we also have additional metadata specifying the poses of the camera and the lights.}
\label{fig:json_structure}
\end{figure}

\begin{table}[t]
\centering
\caption{Factor primitives for various object properties in SVIB-dSprites.}
\vspace{2mm}
\label{tab:factor_primitives_dsprites}
\begin{tabular}{ccc}
\toprule
Shape & Color (RGB) & Size \\
\midrule
\texttt{circle} & \texttt{(0, 255, 0)} & \texttt{0.125} \\
\texttt{triangle} & \texttt{(255, 0, 255)} & \texttt{0.225} \\
\texttt{square} & \texttt{(0, 127, 255)} & \texttt{0.325} \\
\texttt{star\_4} & \texttt{(255, 127, 0)} & \texttt{0.425} \\
\bottomrule
\end{tabular}
\end{table}

\begin{table}[t]
\centering
\caption{Factor primitives for various object properties in SVIB-CLEVR.}
\vspace{2mm}
\label{tab:factor_primitives_clevr}
\begin{tabular}{cccc}
\toprule
Shape & Color (RGB) & Size & Material \\
\midrule
\texttt{SmoothCube\_v2} & \texttt{(255, 0, 0)} & \texttt{1.0} & \texttt{Rubber} \\
\texttt{Sphere} & \texttt{(0, 255, 0)} & \texttt{1.5} & \texttt{MyMetal} \\
\texttt{SmoothCylinder} & \texttt{(0, 0, 255)} & \texttt{2.0} & \\
\texttt{Suzanne} & \texttt{(0, 255, 255)} & & \\
& \texttt{(255, 0, 255)} & & \\
& \texttt{(255, 255, 0)} & & \\
\bottomrule
\end{tabular}
\end{table}

\begin{table}[t]
\centering
\caption{Factor primitives for various object properties in SVIB-CLEVRTex. For materials, we use 8 free textures provided by Poliigon (\url{https://www.poliigon.com})}
\vspace{2mm}
\label{tab:factor_primitives_clevrtex}
\begin{tabular}{cccc}
\toprule
Shape & Size & Material \\
\midrule
\texttt{Cone} & \texttt{1.0} & \texttt{PoliigonBricksFlemishRed001} \\
\texttt{Cube} & \texttt{1.5} & \texttt{PoliigonBricksPaintedWhite001} \\
\texttt{Cylinder} & \texttt{2.0} & \texttt{PoliigonChainmailCopperRoundedThin001} \\
\texttt{Suzanne} & & \texttt{PoliigonFabricDenim003} \\
\texttt{Icosahedron} & & \texttt{PoliigonFabricFleece001} \\
\texttt{NewellTeapot} & & \texttt{PoliigonMetalSpottyDiscoloration001} \\
\texttt{Sphere} & & \texttt{PoliigonRoofTilesTerracotta004} \\
\texttt{Torus} & & \texttt{PoliigonWoodFlooring061} \\
\bottomrule
\end{tabular}
\end{table}

\begin{figure}[t]
    \centering
    \includegraphics[width=\textwidth]{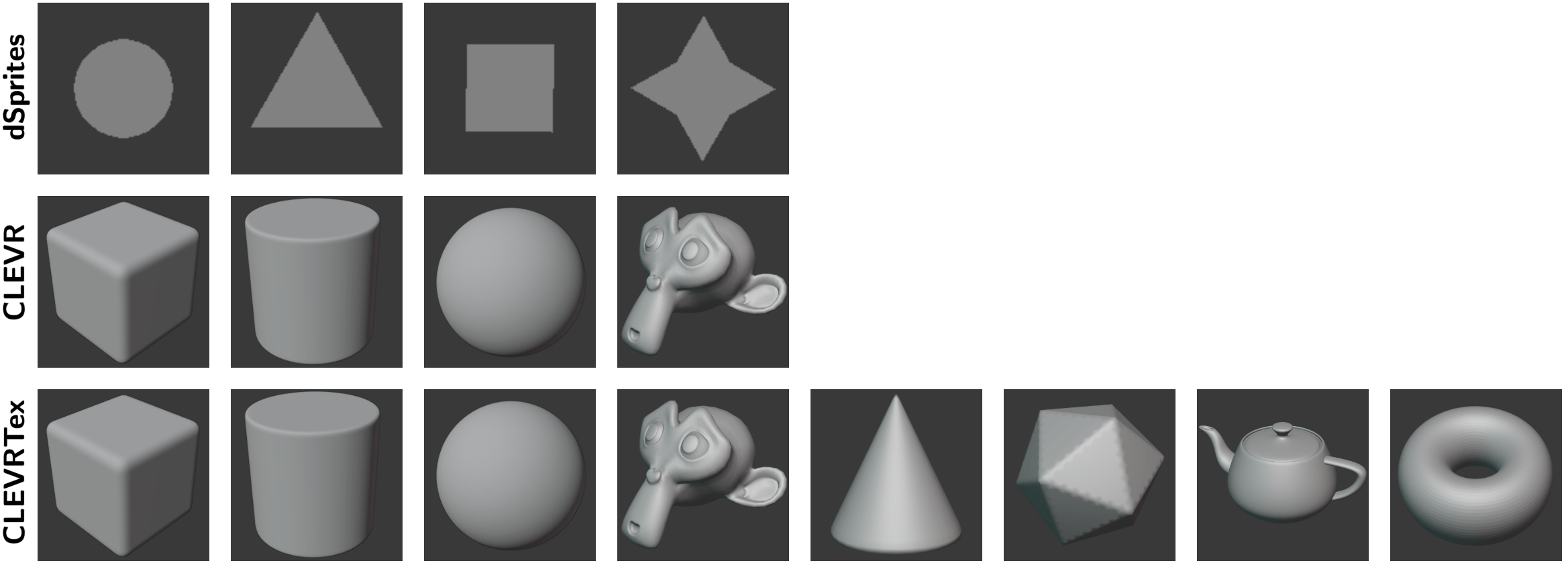}
    \caption{Illustrations of shapes used in various environments of our benchmark. For SVIB-dSprites, the shapes are generated using the Spriteworld API (\url{https://github.com/deepmind/spriteworld}). The shapes used in our benchmark for SVIB-CLEVR and SVIB-CLEVRTex are taken from the Github code (\url{https://github.com/karazijal/clevrtex-generation}) released by Karazija \textit{et al.}\cite{karazija2021clevrtex}.}
    \label{fig:shape_lib}
\end{figure}

\begin{figure}[t]
    \centering
    \includegraphics[width=\textwidth]{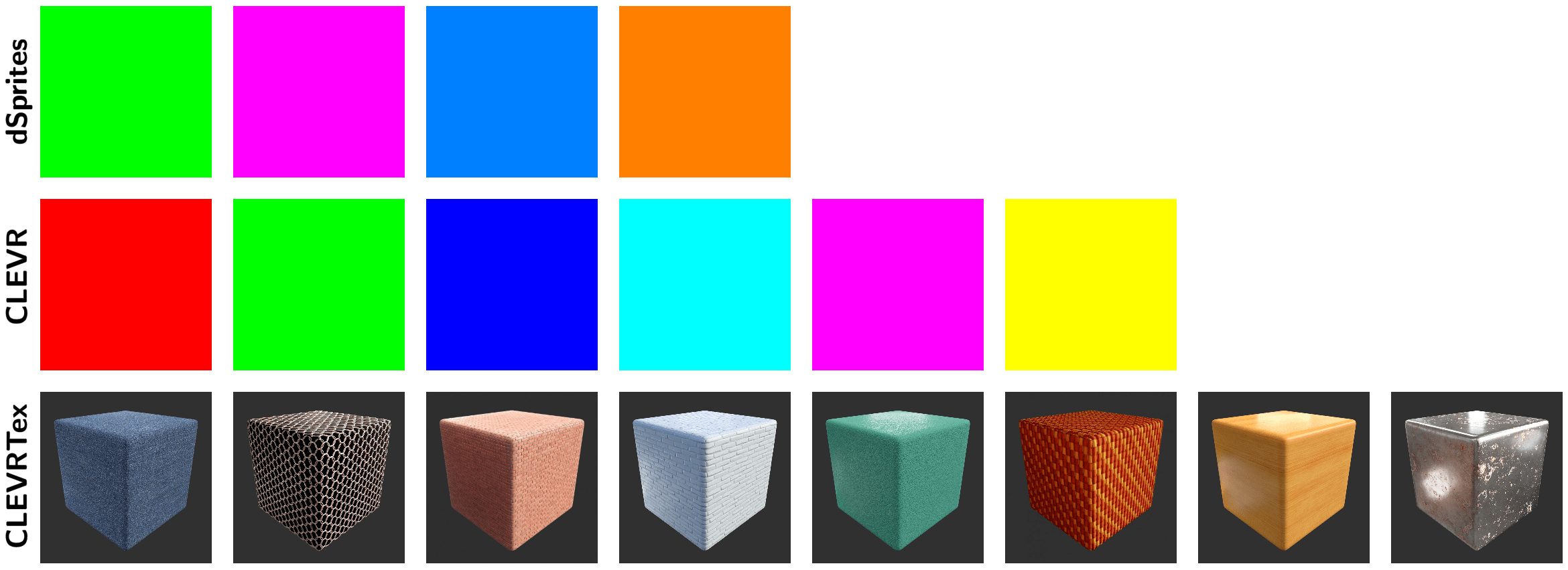}
    \caption{Illustrations of colors and materials used in various environments of our benchmark. For SVIB-CLEVRTex, we use 8 free textures provided by Poliigon (\url{https://www.poliigon.com}) that are also used in the original CLEVRTex data generation code (\url{https://github.com/karazijal/clevrtex-generation}) released by Karazija \textit{et al.}\cite{karazija2021clevrtex}.}
    \label{fig:color_lib}
\end{figure}

\subsubsection*{Question 5: Is there a label or target associated with each instance?} 

Yes, in each instance, the prediction target is a 128$\times$128 RGB image (denoted as the target image) and is provided as the PNG file \texttt{target.png}. 

\subsubsection*{Question 6: Is any information missing from individual instances?}

No, our scenes are not partially observable as we took care to ensure that all objects in our scenes have at least a certain number of visible pixels in the observation images \texttt{source.png} and \texttt{target.png}.

\subsubsection*{Question 7: Are relationships between individual instances made explicit (e.g., users' movie ratings, social network links)?}

No relationships are present between individual instances.

\subsubsection*{Question 8: Are there recommended data splits (e.g., training, development/validation, testing)?}

Yes, for each task, we provide 4 training splits and one OOD test split. The 4 training splits capture different levels of systematic generalization difficulty and correspond to $\alpha$ values $0.0$, $0.2$, $0.4$, and $0.6$. We do not provide a separate validation split, it is completely up to the learner how they want to leverage the training splits for validation e.g., via hold-out validation, $K$-fold cross-validation, leave-one-out validation, etc.

\subsubsection*{Question 9: Are there any errors, sources of noise, or redundancies in the dataset?} No.

\subsubsection*{Question 10: Is the dataset self-contained, or does it link to or otherwise rely on external resources (e.g., websites, tweets, other datasets)?} Yes, the dataset is self-contained and does not require external resources to work with.

\subsubsection*{Question 11: Does the dataset contain data that might be considered confidential (e.g., data that is protected by legal privilege or by doctor-patient confidentiality, data that includes the content of individuals' non-public communications)?}

No.

\subsection{Collection Process}

\subsubsection*{Question 1: How was the data associated with each instance acquired?} The data was procedurally generated using the Spriteworld and Blender APIs.

\subsubsection*{Question 2: What mechanisms or procedures were used to collect the data (e.g., hardware apparatus or sensor, manual human curation, software program, software API)?} The SVIB-dSprites images were generated from the Spriteworld API, the SVIB-CLEVR images using Blender $2.78$, and the SVIB-CLEVRTex using Blender $2.93$. All implementations were done in Python. The Blender processes were run on GPU instead of CPU-only for faster rendering. In the creation of individual splits such as specific training or testing splits, a single instance of a modern Nvidia GPU was enough, demanding less than 25GB of GPU memory.

\subsubsection*{Question 3: If the dataset is a sample from a larger set, what was the sampling strategy (e.g., deterministic, probabilistic with specific sampling probabilities)?} 

To generate a specific split e.g., a specific training or testing split, the combinations of factor values that would be exposed in that split were predefined as discussed in Section \ref{sec:method}. From this predefined set of combinations, a specific combination was selected uniformly at random to instantiate each object within the scene. If a generated scene had an object that was fully occluded by another, or if two objects were too close, such scenes were discarded.

\subsubsection*{Question 4: Who was involved in the data collection process (e.g., students, crowdworkers, contractors) and how were they compensated (e.g., how much were crowdworkers paid)?} The authors of this paper alone were involved.

\subsubsection*{Question 5: Over what timeframe was the data collected?} The data generation process took several months and involved multiple refinement and deliberation steps.

\subsubsection*{Question 6: Were any ethical review processes conducted (e.g., by an institutional review board)?} No.

\subsection{Preprocessing / Cleaning / Labeling}

\subsubsection*{Question 1: Was any preprocessing/cleaning/labeling of the data done (e.g., discretization or bucketing, tokenization, part-of-speech tagging, SIFT feature extraction, removal of instances, processing of missing values)?} No. Since our data was procedurally generated, there was no raw data collected or used. As such, there was nothing to preprocess or clean.

\subsubsection*{Question 2: Was the raw data saved in addition to the preprocessed/cleaned/labeled data (e.g., to support unanticipated future uses)?}
Not Applicable.

\subsubsection*{Question 3: Is the software that was used to preprocess or clean or label the data available?}
Not Applicable.

\subsection{Uses}

\subsubsection*{Question 1: Has the dataset been used for any tasks already?}

Yes, the dataset has been used in our paper to evaluate various state-of-the-art architectures in terms of their ability to systematically generalize.

\subsubsection*{Question 2: Is there a repository that links to any or all papers or systems that use the dataset?}

We plan to list these on the official website of this benchmark.

\subsubsection*{Question 3: What (other) tasks could the dataset be used for?}
Given that we provide scene metadata and object masks, future explorations might investigate the utility of mask supervision in solving our tasks. That said, we also note that using the scene metadata and object masks is not the intended path to solving our benchmark. 

\subsubsection*{Question 4: Is there anything about the composition of the dataset or the way it was collected and preprocessed/cleaned/labeled that might impact future uses?} No, the dataset was procedurally generated. No real data about people was used. As such, it is unlikely that using our data would cause direct harm. While the intended goal of our benchmark is to spur the development of more capable models, future deployments of such models should be mindful of any potential harms.

\subsubsection*{Question 5: Are there any tasks for which the dataset should not be used?}
We are not aware of tasks that, if performed via our dataset, would lead to direct negative consequences.

\subsection{Distribution}

\subsubsection*{Question 1: Will the dataset be distributed to third parties outside of the entity (e.g., company, institution, organization) on behalf of which the dataset was created?} Yes, the dataset is publicly available at \url{https://systematic-visual-imagination.github.io/}.

\subsubsection*{Question 2: How will the dataset be distributed (e.g., tarball on website, API, Github)?} The dataset is distributed at \url{https://systematic-visual-imagination.github.io/} via Google Drive links.

\subsubsection*{Question 3: When will the dataset be distributed?} The dataset is available now at \url{https://systematic-visual-imagination.github.io/}.

\subsubsection*{Question 4: Will the dataset be distributed under a copyright or other intellectual property (IP) license, and/or under applicable terms of use (ToU)?} The dataset will be released under the most liberal Creative Commons license i.e., CC0. While not mandatory, we do encourage future works to cite our paper when using our benchmark.

\subsubsection*{Question 5: Have any third parties imposed IP-based or other restrictions on the data associated with the instances?}
We are not aware of any IP-based restrictions imposed by third parties on our dataset. 

\subsubsection*{Question 6: Do any export controls or other regulatory restrictions apply to the dataset or to individual instances?} We are not aware of any.

\subsection{Maintenance}

\subsubsection*{Question 1: Who is supporting/hosting/maintaining the dataset?} 

The research lab at KAIST led by Sungjin Ahn will maintain the dataset.

\subsubsection*{Question 2: How can the owner/curator/manager of the dataset be contacted (e.g., email address)?} 

The manager can be contacted via email: \url{sungjin.ahn@kaist.ac.kr}

\subsubsection*{Question 3: Is there an erratum?} If errors are found at a later date, we will provide an erratum on the official website.

\subsubsection*{Question 4: Will the dataset be updated (e.g., to correct labeling errors, add new instances, delete instances)?} Yes, in case there are updates, we will release the new version on our website. The older versions will also remain available.

\subsubsection*{Question 5: If the dataset relates to people, are there applicable limits on the retention of the data associated with the instances (e.g., were individuals in question told that their data would be retained for a fixed period of time and then deleted)?} Not Applicable.

\subsubsection*{Question 6: Will older versions of the dataset continue to be supported/hosted/maintained?} Yes.

\subsubsection*{Question 7: If others want to extend/augment/build on or contribute to the dataset, is there a mechanism for them to do so?} 

Yes, we shall release all our source code, including the code we used to generate the datasets under a highly permissive CC0 license. Others can freely download, modify and create their own variants of the dataset.

\subsection{Author Statement of Responsibility}
We, the authors, confirm that we bear all responsibility in the case of violation of rights and licenses.

\section{Additional Experiment Results}
In this section, we provide additional experimental results that could not be included in the main paper due to the limitation of space.
\subsection{MSE on the Benchmark Tasks}
\begin{figure*}[t]
    \centering
    \includegraphics[width=1.0\textwidth]{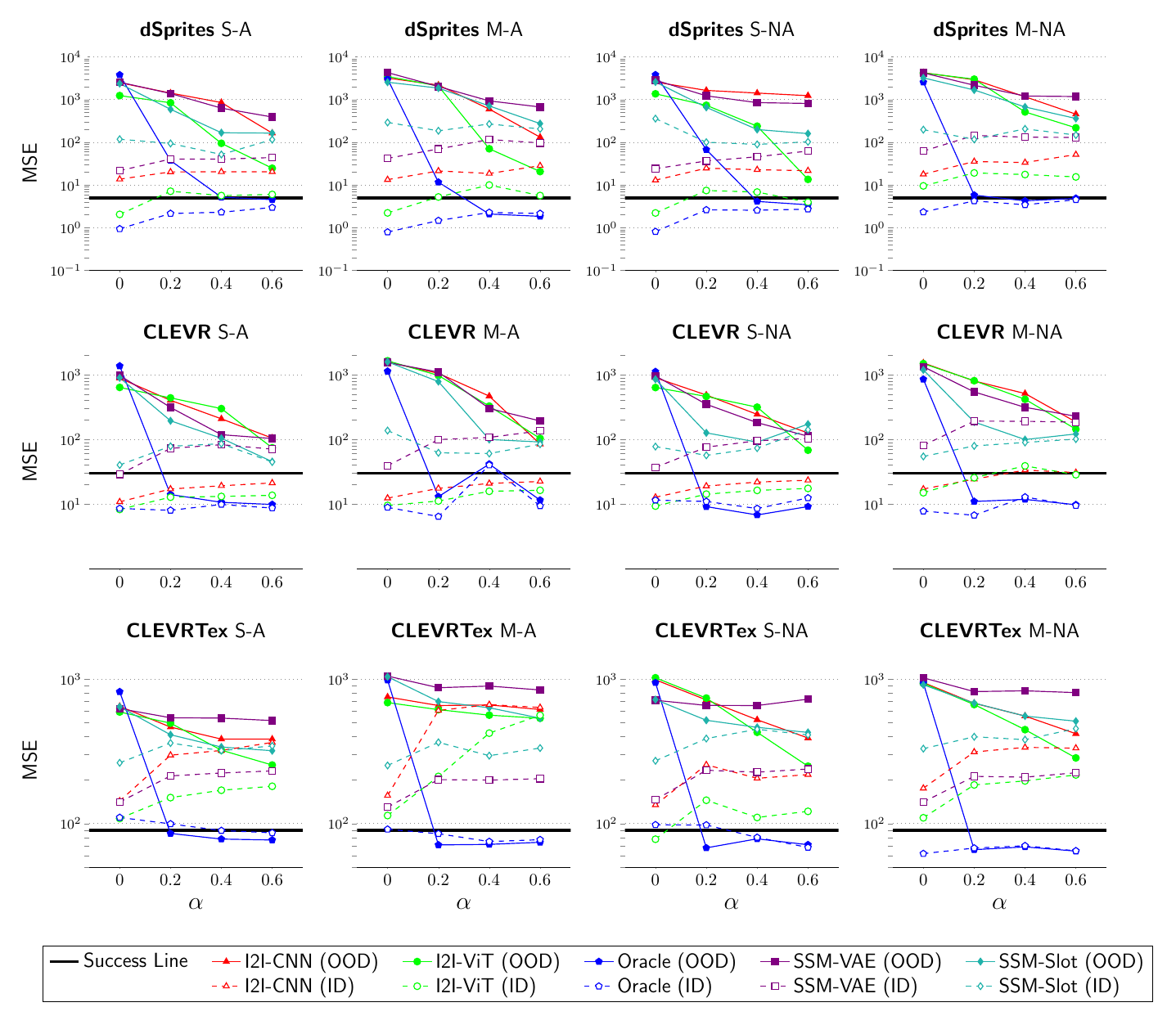}
    \caption{\textbf{MSE Performance.} We report the in-distribution and out-of-distribution MSE for all benchmark tasks. 
    }
    \label{fig:all-reference-mse}
\end{figure*}

In Figure~\ref{fig:all-reference-mse}, we report the in-distribution and out-of-distribution MSE for all benchmark tasks. 

\subsection{Generalization Gap on the Benchmark Tasks}
\begin{figure*}
    \centering
    \includegraphics[width=1.0\textwidth]{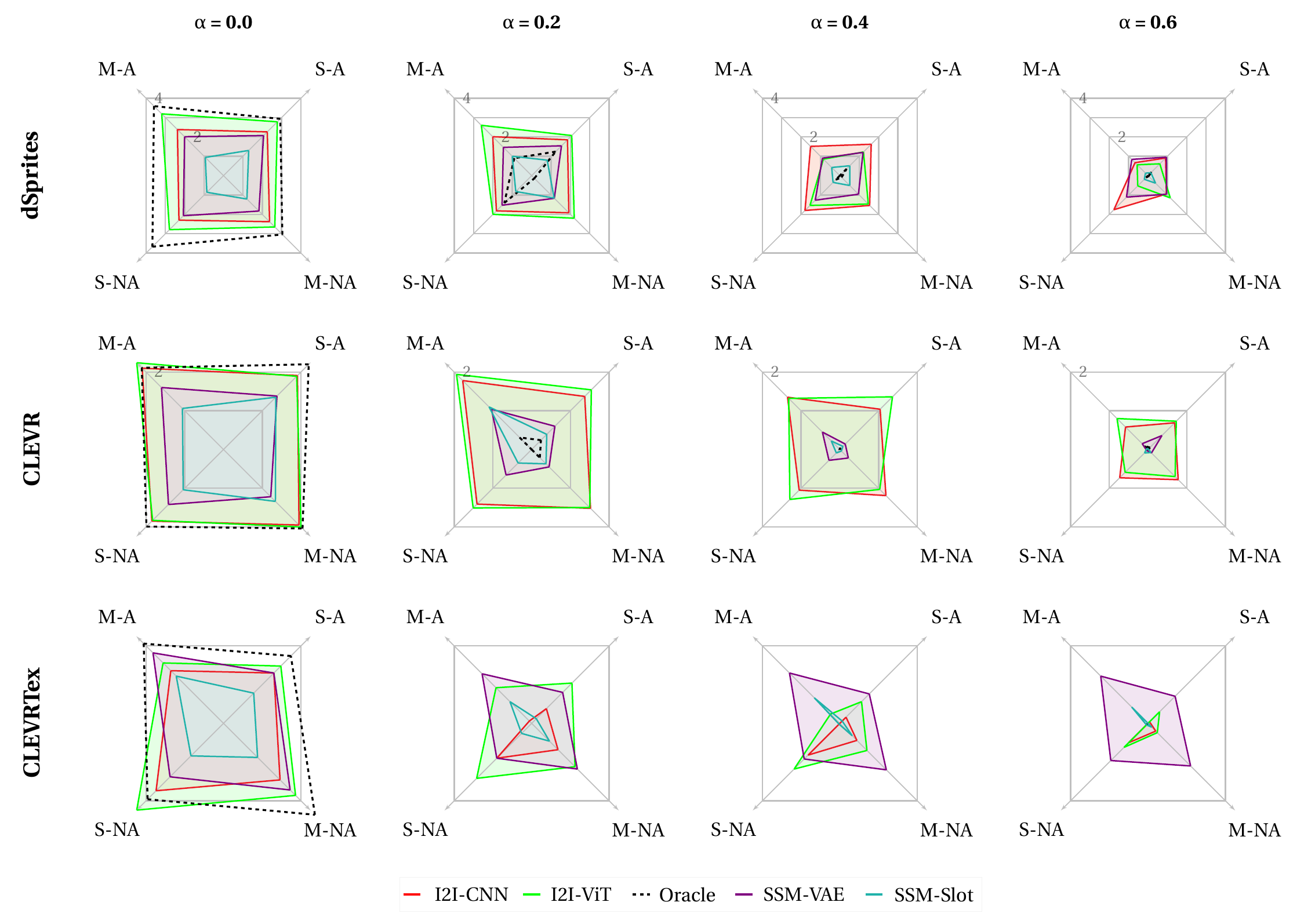}
    \caption{\textbf{Systematic Generalization Gap.} We report the systematic generalization gap for all benchmark tasks.}
    \label{fig:all-reference-sgg}
\end{figure*}

In Figure \ref{fig:all-reference-sgg}, we report the systematic generalization gap for all benchmark tasks. The systematic generalization gap is defined as the difference between the OOD MSE and in-distribution MSE in the log scale.

\subsection{Qualitative Results}
\begin{figure*}
    \centering
    \includegraphics[width=1.0\textwidth]{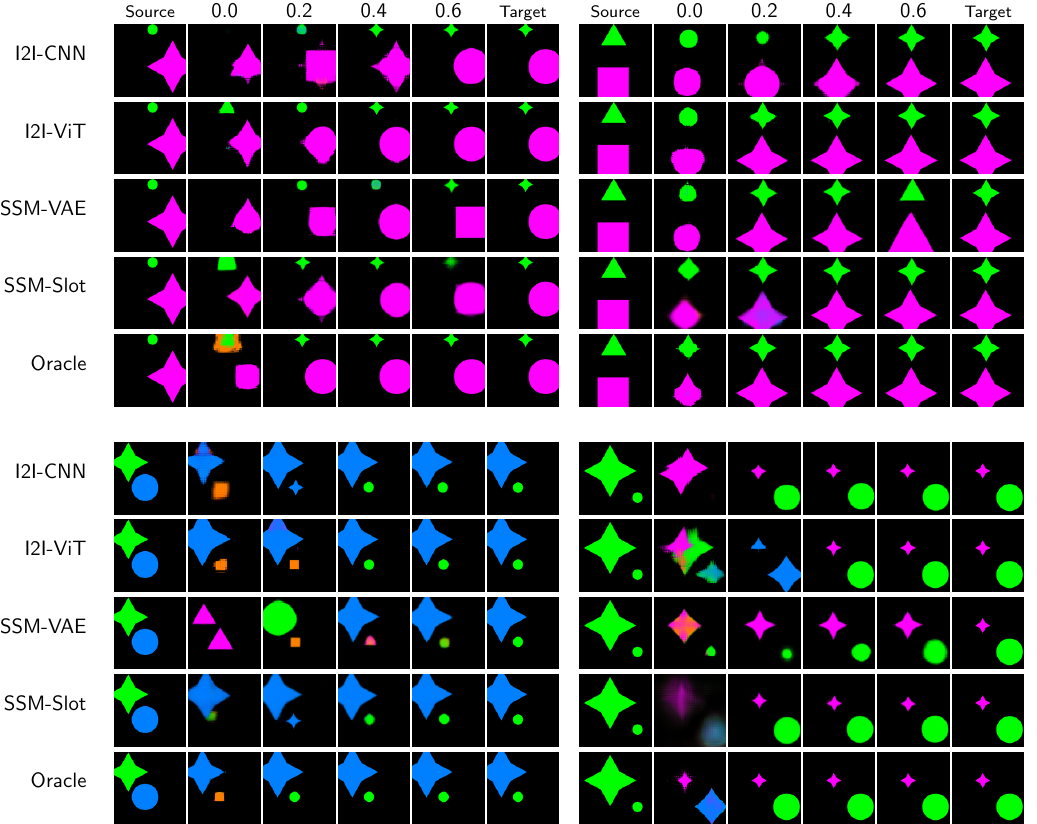}
    \caption{\textbf{Qualitative Results on Benchmark Tasks in SVIB-dSprites.} We visualize the baseline predictions on all 4 tasks. \textit{Top-Left:} Single Atomic. \textit{Top-Right:} Single Non-Atomic. \textit{Bottom-Left:} Multiple Atomic. \textit{Bottom-Right:}  Multiple Non-Atomic.
    }
    \label{fig:dsprites-qualitative}
\end{figure*}

\begin{figure*}
    \centering
    \includegraphics[width=1.0\textwidth]{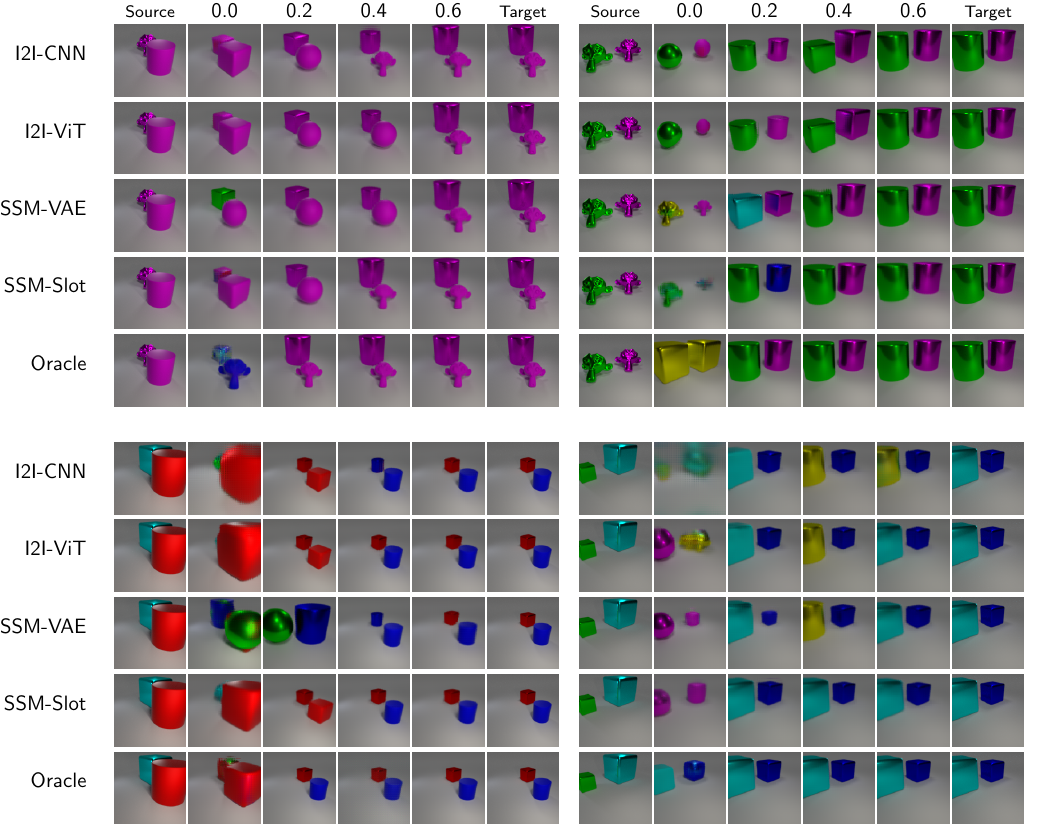}
    \caption{\textbf{Qualitative Results on Benchmark Tasks in SVIB-CLEVR.} We visualize the baseline predictions on all 4 tasks. \textit{Top-Left:} Single Atomic. \textit{Top-Right:} Single Non-Atomic. \textit{Bottom-Left:} Multiple Atomic. \textit{Bottom-Right:}  Multiple Non-Atomic.}
    \label{fig:clevr-qualitative}
\end{figure*}

\begin{figure*}[t]
    \centering
    \includegraphics[width=1.0\textwidth]{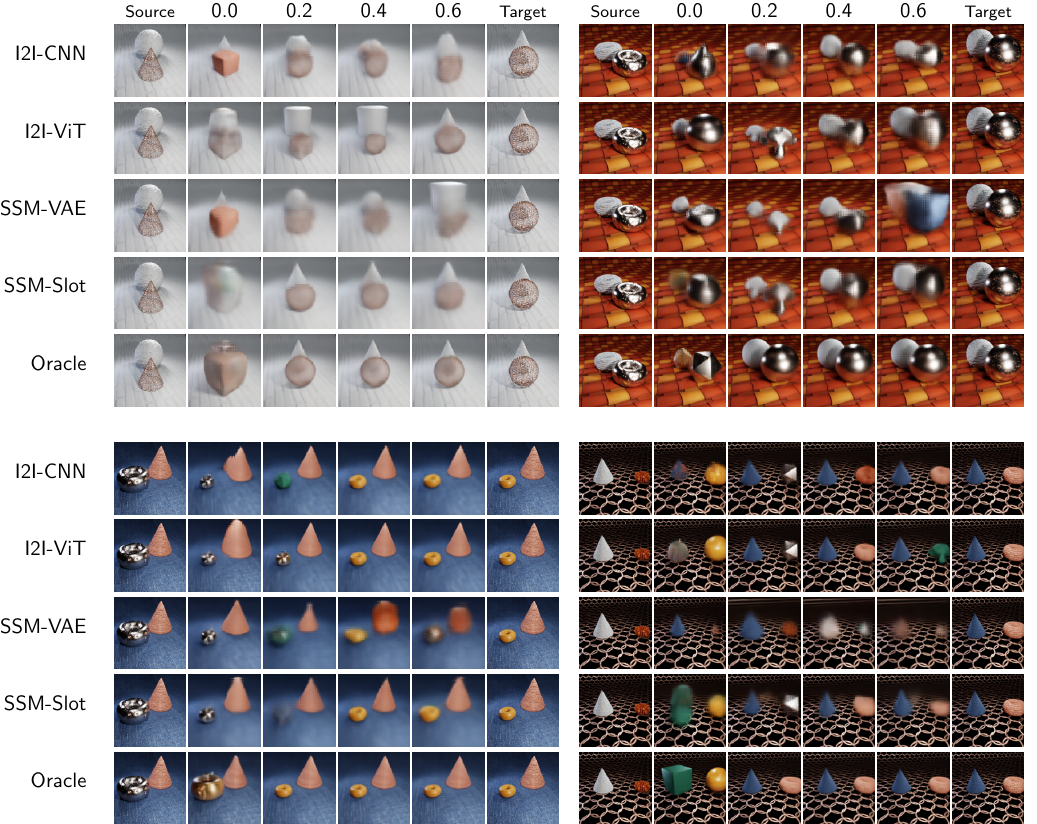}
    \caption{\textbf{Qualitative Results on Benchmark Tasks in SVIB-CLEVRTex.} We visualize the baseline predictions on all 4 tasks. \textit{Top-Left:} Single Atomic. \textit{Top-Right:} Single Non-Atomic. \textit{Bottom-Left:} Multiple Atomic. \textit{Bottom-Right:}  Multiple Non-Atomic.}
    \label{fig:qualitative_clevrtex}
\end{figure*}

In Figures \ref{fig:dsprites-qualitative}, \ref{fig:clevr-qualitative} and \ref{fig:qualitative_clevrtex}, we visualize the predicted images of various baselines on the benchmark tasks. 

\subsection{MSE on the Analysis Tasks}
\begin{figure*}[t]
    \centering
\includegraphics[width=1.0\textwidth]{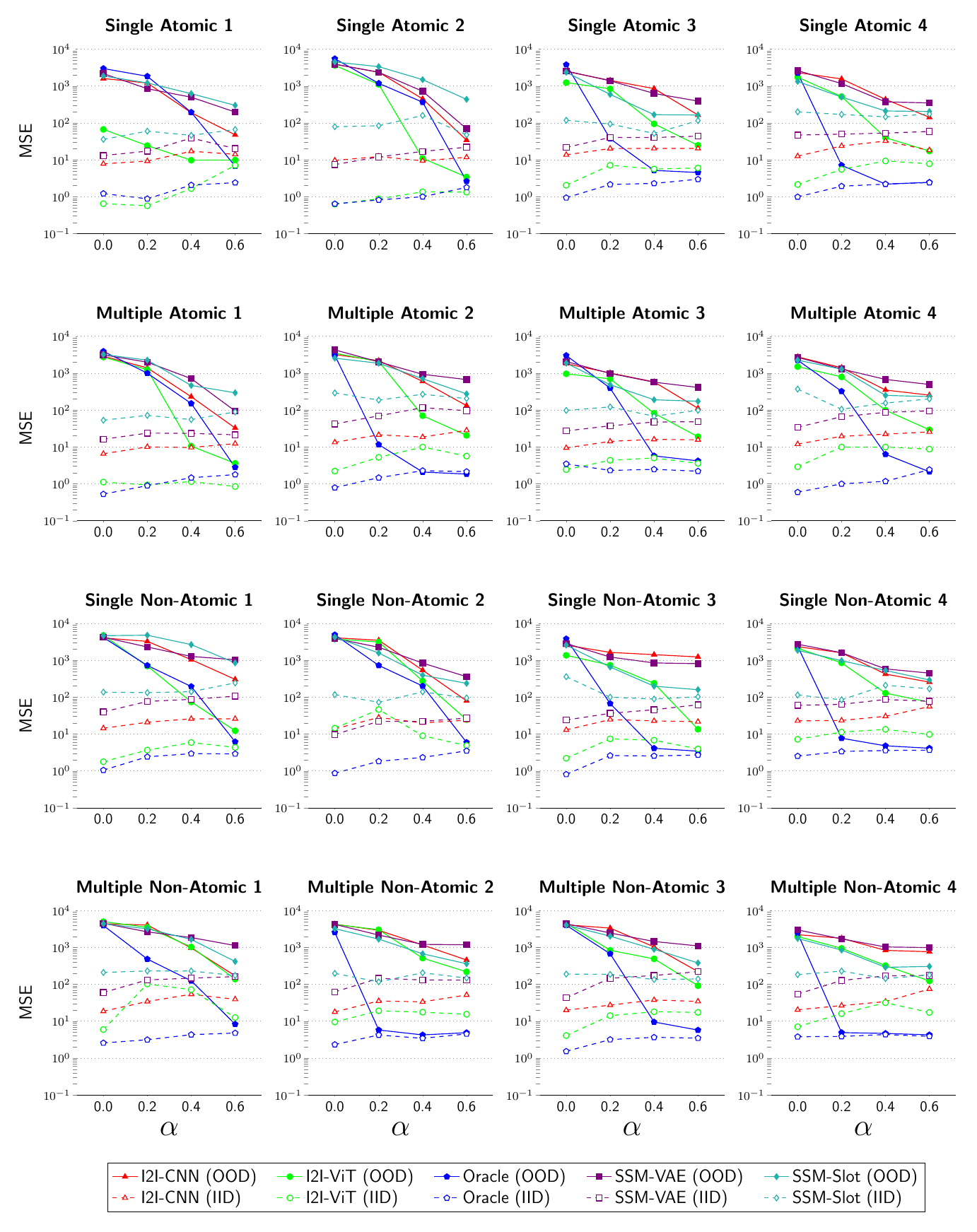}
    \caption{\textbf{MSE Performance on Analysis Tasks.}}
    \label{fig:total-analysis-graph}
\end{figure*}

In Figure \ref{fig:total-analysis-graph}, we plot the MSE performance on the analysis tasks. A detailed description of how the analysis tasks are constructed is provided in Section \ref{sec:analysis_tasks}.

\subsection{Comparison of In-Distribution Performance between Image-to-Image and State-Space Models}
In Figures \ref{fig:all-reference-lpips} and \ref{fig:all-reference-mse}, one observation is that state-space models (SSMs) generally have worse in-distribution MSE than image-to-image models. This can be expected since the image-to-image models minimize the prediction error directly in the image space while state-space models minimize the prediction error in the latent space which is an indirect objective.

\subsection{Factor Prediction}
The SVIB-CLEVRTex environment, which is highly textured data, has a relatively high level of visual complexity. Consequently, within this environment, the importance of visual recognition abilities can significantly increase. Therefore, we conducted an experiment in which we trained a simple 4-layer CNN encoder to take SVIB-CLEVRTex images as input and predict GT labels to ascertain whether $128\times 128$ image resolution can provide sufficient information in this setting.

\begin{table}[t]
\centering
\caption{\textbf{Accuracy of Ground-Truth Factor Prediction in SVIB-CLEVRTex}. We perform a factor prediction task where we take SVIB-CLEVRTex images as input and predict the factors for the scene. We indicate these by $\text{Shape}_{i}$, $\text{Size}_{i}$, $\text{Mat}_{i}$, where $i=1,2$ denotes object index in the scene based on closeness to the camera. We also predict the background material denoted with $\text{Mat}_\text{bg}$.}
\vspace{2mm}
\resizebox{\textwidth}{!}{
\begin{tabular}{l C{1.5cm} C{1.5cm} C{1.5cm} C{1.5cm} C{1.5cm} C{1.5cm} C{1.5cm}}
\toprule
\textbf{Test Accuracy}      &   $\text{Shape}_{1}$   &   $\text{Shape}_{2}$   &  $\text{Size}_{1}$ &   $\text{Size}_{2}$    &   $\text{Mat}_{1}$ &   $\text{Mat}_{2}$ &   $\text{Mat}_\text{bg}$    \\
\midrule
Top-1 Acc. (\%)             &   92.81       &   88.39       &   94.99   &   93.46       &   95.67   &   92.98   &   99.99       \\
\bottomrule
\end{tabular}
}
\label{table:gt-prediction}
\end{table}

\section{Additional Related Work}

\makeatletter
\def\adl@drawiv#1#2#3{%
        \hskip.5\tabcolsep
        \xleaders#3{#2.5\@tempdimb #1{1}#2.5\@tempdimb}%
                #2\z@ plus1fil minus1fil\relax
        \hskip.5\tabcolsep}
\newcommand{\cdashlinelr}[1]{%
  \noalign{\vskip\aboverulesep
           \global\let\@dashdrawstore\adl@draw
           \global\let\adl@draw\adl@drawiv}
  \cdashline{#1}
  \noalign{\global\let\adl@draw\@dashdrawstore
           \vskip\belowrulesep}}
\makeatother

\newcolumntype{g}{>{\columncolor[gray]{0.9}}l}
\newcolumntype{G}{>{\columncolor[gray]{0.8}}l}

\newcolumntype{H}{>{\columncolor[rgb]{0.9,0.9,0.9}}c}
\newcolumntype{h}{>{\columncolor[rgb]{0.9,0.9,0.9}}c}

\begin{table}[t]
\centering
\caption{\textbf{Limitations of Existing Studies:} The table contrasts our proposed Systematic Visual Imagination Benchmark (SVIB) with the existing studies. We note that existing studies do not offer a benchmark for evaluating systematic perception ability in the image domain.}
\vspace{2mm}
\resizebox{\textwidth}{!}{
\begin{tabular}{Ggp{3.8cm}p{4.5cm}p{3.8cm}p{1cm}}  
    \toprule
    \multicolumn{2}{g}{\textbf{Study}}                    & \multicolumn{1}{l}{\textbf{Task Modality}} & \multicolumn{1}{l}{\textbf{Systematic Perception}}   & \multicolumn{1}{l}{\textbf{Perceptual Complexity}}\\
    \midrule
                                            & SCAN~\citep{scan_lake}          & Text $\rightarrow$ Text           & Not Applicable                       & Not Applicable              \\
    \multirow{-2}{*}{Language}              & gSCAN~\citep{ruis2020benchmark}         & Image + Text $\rightarrow$ Text    & \cmark\;                           & \tmark\;(Toy Multi-Object) \\
    \cdashlinelr{1-6}
                                            & Xu \emph{et al.}~\cite{xu2022compositional}      & Image $\rightarrow$ Latent                  & \cmark\;                           & \tmark\;(Toy Single-Object)\\
    \multirow{-2}{*}{Disentanglement}      & Montero \emph{et al.}~\cite{montero2021role} & Image $\rightarrow$ Latent                  & \cmark\;                           & \tmark\;(Toy Single-Object)\\
    \cdashlinelr{1-6}
                                            & ARC~\citep{chollet2019measure}           & Image $\rightarrow$ Image         & \xmark\;(Non-Systematic)                       & \xmark\;(Very Low Resolution) \\
    \multirow{-2}{*}{Visual Reasoning}             & Sort-of-ARC~\citep{assouel2022object}   & Image $\rightarrow$ Image         & \xmark\;(In-Distribution)                      & \xmark\;(Very Low Resolution) \\
    \midrule
    \multicolumn{2}{g}{\textbf{SVIB (Ours)}}                & Image $\rightarrow$ Image         & \cmark\;                           & \cmark\;(Realistic Multi-Object)  \\
    \bottomrule
    \noalign{\smallskip}\noalign{\smallskip}
\end{tabular}
}
\label{table:dataset-comparison}

\end{table}

\textbf{Benchmarks for Video Prediction.} 
In the realm of video prediction, there are several real-world benchmarks \citep{kitti, humans36m, cityscapes, robonet}. However, these cannot be used to study systematic visual imagination since, being real-world datasets, do not provide control over factor combinations and their demarcation between training and testing. 
Although synthetic benchmarks provide more control over such data-generating factors: \citep{physion, movingmnist, shapestacks, yi2019clevrer, intphys, cophy, gswm}, such benchmarks have only so far focused on in-distribution video prediction and do not focus on systematic out-of-distribution evaluation.

\textbf{Benchmarks for Generalization in RL.} Although OGRE and NovPhy \citep{ogre, novphy} provide a platform to study out-of-distribution objects, these lack well-defined primitives and systematic evaluation unlike ours. 
\citep{Ke2021SystematicEO} focuses on causal induction rather than systematic perception unlike ours.
\citep{powderworld} also provides a platform to study out-of-distribution environments, however, its scene dynamics involve interactions between low-level powder particles rather than high-level abstractions, unlike ours. Yet, all of these are RL benchmarks meaning that they do not provide an isolated way of studying the world modeling itself which is possible in our benchmark.

\textbf{Large-Scale Pretraining and Prompting.} Large-scale pretraining of vision models followed by prompting has shown remarkable zero-shot abilities. Some of these are focused on performing classification or class-conditional image generation via prompting \cite{Jia2022VisualPT, Wu2022GenerativeVP} unlike ours which focuses on image-to-image generation tasks. A more recent line focuses on image-to-image prediction via prompting \citep{painter, visualprompting, seggpt, Sohn2023LearningDP}. A large-scale pre-trained image-to-image model can indeed have capabilities as a world model given appropriate in-context prompts.
However, since large-scale pretraining is done on real datasets without access to the underlying factors, it is difficult to quantify how well the pre-trained model generalizes. On the other hand, in our benchmark, the underlying factors are known and generalization ability can be clearly assessed in terms of $\alpha$ required for task success. Furthermore, large-scale pretraining is an expensive endeavor and which makes it difficult to quickly test and analyze models. On the other hand, with our simple and lightweight benchmark, it is possible to test a model with less than a single modern GPU within a 2-day training timeframe. There have been some studies on the compositional generalization ability of large models, however, to our knowledge, these are confined to the text domain \cite{Levy2022DiverseDI, An2023HowDI}.

\section{Details of the Benchmark}

\subsection{Rule Definitions}
In this section, we will provide a pseudo-code for all the rules proposed in this benchmark. We first describe notations that we use in our pseudo-code and then proceed to describe the rules.

\textbf{Symmetricity.} Our rules are designed to be applied symmetrically to both objects. As such, we describe the rule definitions relative to just one of the objects that we denote as \texttt{self}. We will denote the other object as \texttt{other}. We will access a factor of an object using square brackets e.g., \texttt{other[`color']} shall denote the color of the \texttt{other} object.

\textbf{Integer Factor Value.} In the following descriptions, we will assume that the value of a factor is represented as an integer. For instance, the color of an object \texttt{self[`color']} can take a value in $0, 1, \ldots, \texttt{num\_colors}-1$, where \texttt{num\_colors} is the total number of color primitives in the color vocabulary. Similarly, the shape of an object \texttt{self[`shape']} can take a value in $0, 1, \ldots, \texttt{num\_shapes}-1$, where \texttt{num\_shapes} is the total number of shape primitives in the shape vocabulary.

\subsubsection{Benchmark Rules}
\label{ax:reference-rule-defs}
We define four benchmark tasks for each subset: SVIB-dSprites, SVIB-CLEVR, and SVIB-CLEVRTex. These tasks align with the four rule categories outlined in Section \ref{sec:method}. 

\textbf{Single Atomic.} In this task, the transformation is executed by swapping the shapes of the two objects in the input image. For brevity, we sometimes call it the \textit{Shape-Swap} task and denote it as S-A. 
\begin{align*}
    \texttt{self[`shape']} &\leftarrow \texttt{other[`shape']}
\end{align*}

\textbf{Single Non-Atomic.} This task involves a transformation where the shape of each object is updated based on the shapes of both objects in the input image, as determined by a lookup table. For brevity, we sometimes denote this task as S-NA.
\begin{align*}
    \texttt{self[`shape']} &\leftarrow (\texttt{self[`shape']} + \texttt{other[`shape']}) \mod \texttt{num\_shapes}
\end{align*}

\textbf{Multiple Atomic.} In this task, the transformation involves simultaneous updates to the color and size of each object. The new color is determined by the object's own shape, and the new size is determined by the color of the other object. Both determinations are made by a lookup table. For brevity, we sometimes denote this task as M-A.
\begin{align*}
\texttt{self[`color']} &\leftarrow \texttt{self[`shape']} \mod \texttt{num\_colors}\\
\texttt{self[`size']} &\leftarrow \texttt{other[`color']} \mod \texttt{num\_sizes}
\end{align*}

\textbf{Multiple Non-Atomic.} In this task, the transformation involves simultaneous updates to the color and size of each object. The new color is determined by the object's own shape and quadrant, and the new size is determined by the other object's color and quadrant. Both determinations are made by a lookup table. For brevity, we sometimes denote this task as M-NA.
\begin{align*}
\texttt{self[`color']}  &\leftarrow \texttt{(self[`shape']} \\
                        &\qquad \texttt{+ quadrant(self[`position']))} \mod \texttt{num\_colors}\\
\texttt{self['size']}   &\leftarrow \texttt{(other[`color']} \\
                        &\qquad \texttt{+ quadrant(self[`position']))} \mod \texttt{num\_sizes}
\end{align*}

For SVIB-CLEVRTex, the color factor indicates texture.

\subsubsection{Analysis Tasks}
\label{sec:analysis_tasks}
\begin{figure*}[t]
    \centering
\includegraphics[width=1.0\textwidth]{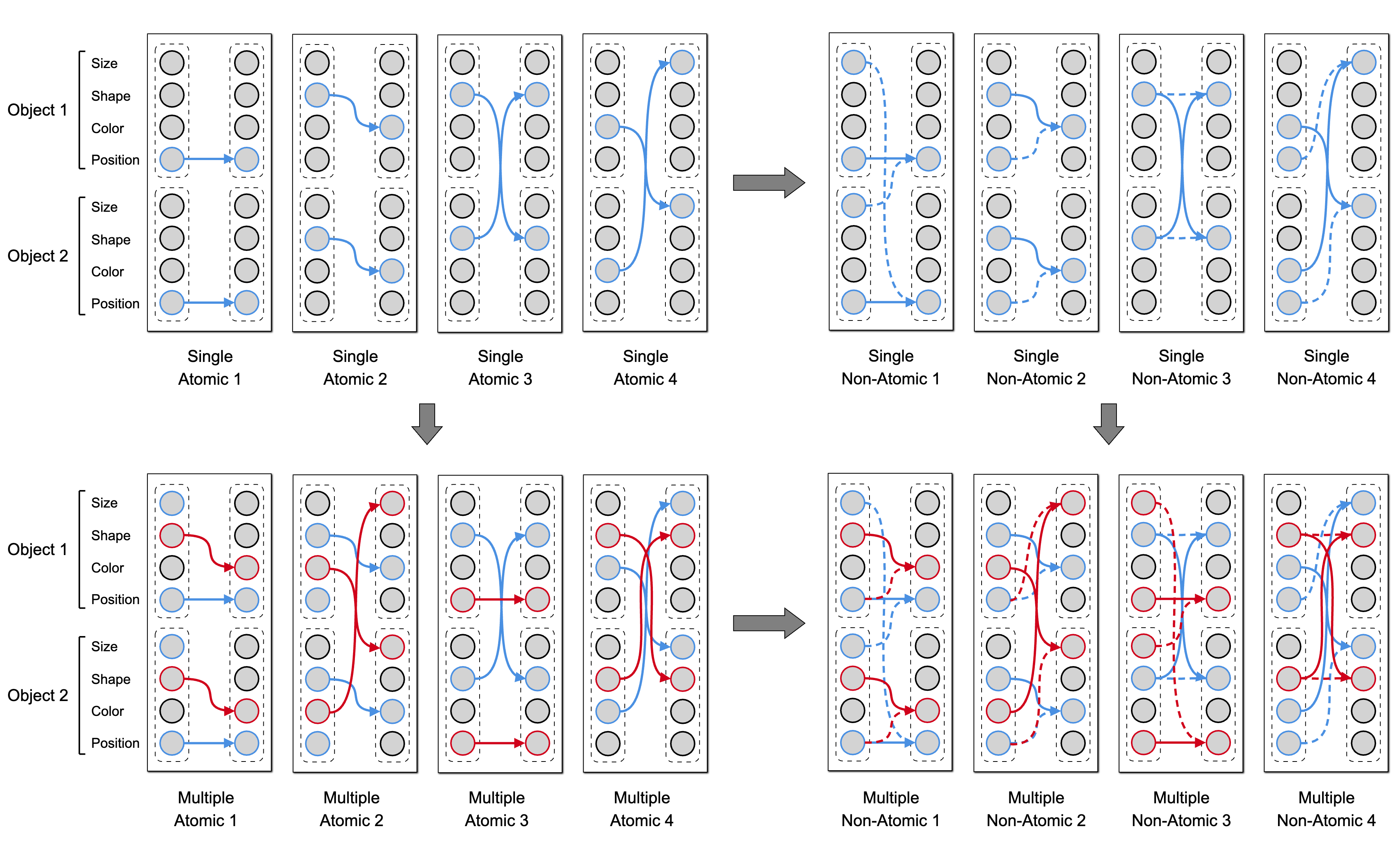}
    \caption{\textbf{Design of Analysis Set}. 
    In this figure, we illustrate the 16 analysis rules using causal graphs, where nodes represent factors. The value of each target node is determined by the source nodes connected by incoming edges.  If a node has just one incoming edge, the integer value of the source factor is directly assigned to the target factor, taking into account the cardinality of the target factor's vocabulary through modulo operation. When a node has two incoming edges, the integer values from both source factors are summed and then assigned to the target factor, subject to modulo based on the cardinality of the target factor's vocabulary. For example, in the \textit{"Single Non-Atomic 4"} rule, two incoming arrows to the 'size' factor signify that its modified value is calculated by adding the quadrant/position of the same object to the color index of the other object, followed by a modulo operation with respect to the cardinality of object size vocabulary. All 16 rules depicted in this diagram can be understood in a similar manner.}
    \label{fig:dsprites-analysis-set-design}
\end{figure*}

We design 16 analysis tasks, a broad pool of tasks from which we eventually choose the rules to construct the final benchmark. The analysis rules in these tasks are designed to be broad-based by keeping the following points in mind:
\enums{
\item All rules taken together should cover all factors. 
\item All rules taken together should cover various types of factor interactions: no interaction, interactions between factors of the same object, and interactions between factors of different objects.
}
With these design considerations we design the rules as shown in Figure~\ref{fig:dsprites-analysis-set-design}. For this, we first create four atomic rules. We then construct rules with greater complexity by incrementally adding factors and parent edges to the causal graph. By analyzing the model performance on these rules in the dSprites environment, we choose 4 rules---one per each rule complexity category---that will be used to create the final benchmark tasks. The rules for the final benchmark tasks are chosen by identifying the rules for which Oracle can solve the tasks well while the other baselines struggle. This is done to balance the difficulty and solvability of our benchmark. The fact that Oracle can solve the tasks shows that our tasks are indeed solvable and do not pose an insurmountable challenge to the community. At the same time, we also ensure that the tasks we choose are difficult to solve for the current models. Note that it is expensive to run 5 baselines on 16 tasks (totaling 80 experiments) for all environments. Therefore, we perform the analysis experiments only in the dSprites environment. The 4 finalized rules are then used for SVIB-CLEVR and SVIB-CLEVRTex as well. When adopting the finalized rules to SVIB-CLEVRTex, we interpret the color factor of the SVIB-dSprites environment as the texture factor of the SVIB-CLEVRTex environment. 

\subsection{Core Combinations}
\label{ax:core}
To show each primitive in the visual vocabularies individually during training, we create a set of core combinations. This is the smallest set that contains all visual primitives at least once. Given the factor vocabularies  $\mathcal{V}_1, \ldots, \mathcal{V}_M$, we construct the core combinations as described in Algorithm \ref{algo:core-combinations}.
\begin{algorithm}[t]
\caption{Collecting Core Combinations}
\label{algo:core-combinations}
\begin{algorithmic}[1]
\STATE \textbf{Input:} Vocabularies \(\mathcal{V}_1, \ldots, \mathcal{V}_M\)
\STATE \textbf{Output:} Set \(\cC\) containing core combinations, each combination represented as tuple.
\STATE
\STATE \textbf{Initialize} an empty set \(\cC\)
\STATE \(N \gets \max_{m=1}^{M} |\mathcal{V}_m|\)
\STATE
\STATE \textbf{for} \(i = 0\) \textbf{to} \(N-1\) \textbf{do}
\STATE \quad \textbf{Initialize} an empty tuple \(T\)
\STATE \quad \textbf{for} \(m = 1\) \textbf{to} \(M\) \textbf{do}
\STATE \quad \quad \textbf{Append} \(\mathcal{V}_{m, i \text{ modulo } |\cV_m|}\) to \(T\)
\STATE \quad \textbf{end for}
\STATE \quad \textbf{Add} \(T\) to \(\cC\)
\STATE \textbf{end for}
\STATE
\STATE \textbf{return} \(\cC\)
\end{algorithmic}
\end{algorithm}

\section{Details of the Baselines}

\begin{table}[t]
\caption{\textbf{Baseline Hyperparameters.} In this table, we provide the hyperparameters for all baselines evaluated in our experiments.}
\vspace{2mm}
\centering
\small
\begin{tabular}{@{}llccc@{}}
\toprule
                &                   & \multicolumn{3}{c}{Benchmark Subset}                          \\ \cmidrule(l){3-5} 
Module          &   Hyperparameter  & SVIB-dSprites      & SVIB-CLEVR     & SVIB-CLEVRTex                 \\ \midrule
General         &   Batch Size      & 32            & 40        & 40                       \\
                &   Training Steps  & 160K          & 160K      & 160K                     \\
                \midrule
CNN Encoder     &   Kernel Size     & 5             & 5         & 5                         \\
                &   Stride          & 2             & 2         & 2                         \\
                &   Padding         & 2             & 2         & 2                         \\
                &   Hidden Size     & 64            & 64        & 64                        \\
                &   Learning Rate   & 0.0001        & 0.0001    & 0.0001                    \\
                \midrule
ViT Encoder     & \# Encoder Blocks & 8             & 8         & 8                         \\
                & \# Encoder Heads  & 8             & 8         & 8                         \\
                &   Hidden Size     & 192           & 192       & 192                       \\
                &   Dropout         & 0.1           & 0.1       & 0.1                       \\
                &   Learning Rate   & 0.0001        & 0.0001    & 0.0001                    \\
                \midrule
SSM-VAE         & \# Dynamic Blocks & 4             & 4         & 4                         \\
                &   VAE Latents     & 64            & 64        & 64                        \\
                &   VAE $\beta$       & 1.0           & 1.0       & 1.0                       \\
                &   VAE $\sigma$      & 0.01          & 0.01      & 0.01                      \\
                &   Learning Rate   & 0.0003        & 0.0003    & 0.0003                    \\
                \midrule
SSM-Slot        & \# Dynamic Blocks & 4             & 4         & 4                         \\
                & \# Dynamic Heads  & 4             & 4         & 4                         \\
                &   \# Slots        & 4             & 4         & 4                         \\
                &   Slot Size       & 64            & 64        & 64                        \\
                &   \# Refinement Iterations   & 3             & 3         & 3                         \\
                &   Learning Rate   & 0.0002        & 0.0002    & 0.0002                    \\
                \midrule
Transformer Decoder
                & \# Decoder Blocks  & 8             & 8         & 8                         \\
                & \# Decoder Heads   & 4             & 4         & 4                         \\
                & Patch Size        & $4\times4$ pixels & $4\times4$ pixels & $4\times4$ pixels \\
                & Hidden Size       & 192           & 192       & 192                       \\
                & Dropout           & 0.1           & 0.1       & 0.1                       \\
                & Learning Rate     & 0.0003        & 0.0003    & 0.0003                    \\
                \bottomrule
\end{tabular}
\label{tab:hyperparams}
\end{table}

In this section, we provide the details of the implementation of the baselines. In Table~\ref{tab:hyperparams}, we report the hyperparameter for modules used in our baselines. 

\subsection{Image-to-Image Models}
The image-to-image models consist of an encoder and a decoder.
\begin{align*}
\bee = \text{Encoder}_\ta(\bx) && \Longrightarrow && \hat{\by} = \text{Decoder}_\gamma(\bee)
\end{align*}
We implement two variants for the encoder: CNN and ViT. To implement the decoder, we adopt a transformer decoder. The complete model is trained in an end-to-end manner by minimizing the mean squared error (MSE) between the predicted image and the target image, i.e., $\cL(\ta, \gamma) = ||\hat{\by} - \by ||^2$.

\subsection{State-Space Models}
To train the modules of the state-space models, we adopt the following three-stage approach:
\enums{
\item In the first stage, we train the encoder $\text{Encoder}_\phi$ network. For this, we use the combined dataset of input and target images of a given task.
\item In the second stage, we freeze the $\text{Encoder}_\phi$ and train only the dynamics model $\text{Dynamics}_\ta$ via a simple latent-level MSE loss: $\cL(\ta) = ||\text{Dynamics}_\ta(\bz_\bx) - \bz_\by||^2$. This dynamics model is implemented as a 4-layer transformer.
\item In the third stage, we freeze the encoder $\phi$ and the dynamics model $\ta$ and train a \textit{probe} parametrized by $\gamma$. The probe takes the latent $\bz_\by$ predicted by the dynamics model and decodes it to render the target image $\hat{\by}$. We deliberately implement the probe as a transformer decoder using the same implementation that was also used for the Image-to-Image models and the Oracle to perform a fair comparison of performance across baselines.
}

\subsubsection{SSM-VAE}

We train the VAE on the task images using its standard auto-encoding objective. After training the VAE, we use the mean of the posterior network as the $\text{Encoder}_\phi(\cdot)$ to acquire latent representations of task images. Given the latent representations, we train a dynamics model to predict the target latent given the input latent by minimizing a latent-level MSE objective.

\subsubsection{SSM-Slot}

We train SAVi in a fully unsupervised manner, by considering input and target images $(\bx, \by)$ from a task as a 2-frame video. 
After training the SAVi, we utilize the encoder portion to obtain the aligned slot representations for the task images. Given the slot representations, we train a dynamics model to predict the slots for the target image given the slots for the input image by minimizing a slot-level MSE objective.

\subsection{Oracle}
\label{ax:oracle}
Oracle is designed to bypass the perception step by directly taking the ground truth scene factors as input. Oracle consists of an encoder that maps the ground truth scene factors to a representation comprised of a set of vectors
$\bee \in \mathbb{R}^{(NM+G)\times D}$, where $N$ is the number of objects, $M$ is the number of factors per object and $G$ is the number of global scene factors such as background texture, lighting, camera, etc. $D$ is the size of each embedding. 
In $\bee$, the categorical factors are represented via learned embeddings while the float factors are represented as sine-cosine embeddings. With this design, Oracle has the perfect systematic generalization ability for perception since it always receives a correctly factored multi-vector representation as the input, regardless of what the input image is.
Finally, a transformer decoder decodes $\bee$ to predict the target image. 
\begin{align*}
    \bee = \text{Encoder}_\phi(\bs_\bx) && \Longrightarrow && \hat{\by} = \text{TransformerDecoder}_\gamma(\bee)
\end{align*}

\subsection{Transformer Decoder}
\label{sec:transformer_decoder}

In previous sections, we refer to the use of a transformer decoder for 1) predicting the target image in the image-to-image models and 2) as a probe to predict the target image given the predicted target state by the SSMs. In this section, we describe the implementation of this transformer decoder. 

Given an input representation, the transformer decoder predicts the patches of the target image. With $H_\text{patch} \times W_\text{patch}$ as the dimensions of a patch, an $H\times W$ image can be seen as a collection of $L = \frac{H\times W}{H_\text{patch} \times W_\text{patch}}$ patches. The transformer is given a collection of positional embeddings as input---one for each of the $L$ patches. Within the transformer, a layer of transformer performs self-attention between these patches, cross-attention on the given input representation to decode, followed by a standard MLP layer. All these steps are done with residual connections. The output of the transformer is $L$ embeddings. Each of these embeddings is mapped and reshaped using a linear layer to generate each patch of size $H_\text{patch} \times W_\text{patch} \times C$, where $C$ is the number of channels ($C=3$ in our case for RGB images). 

A similar decoder design has also been employed in models such as MAE and OSRT \cite{mae, osrt}. Compared to CNN decoders which have local receptive fields, the transformer enables much longer range interactions via the attention layers \cite{imagetransformer, imagegpt}.

\section{Discussion}

\textbf{Balancing Difficulty and Solvability.} Although our results show that our benchmark is unsolved, yet, while designing this benchmark, we took caution to avoid posing an excessively difficult or impossible challenge to the community. The solvability of our benchmark is confirmed by the success of Oracle on our benchmark. Such solvability is a crucial aspect of our benchmark which sets us apart from other benchmarks like ARC \citep{chollet2019measure} which eventually turned out too difficult.

\textbf{Alpha Efficiency.} 
A model that can generalize with smaller $\alpha$ can be considered superior to a model that requires a much larger $\alpha$ value to generalize. By providing the ability to measure the required $\alpha$, our benchmark provides a novel measurable objective for making new modeling advances. For instance, although ViT can generalize on SVIB-dSprites for $\alpha\geq 0.4$, Oracle can do it with smaller $\alpha\geq 0.2$, making it superior to ViT.

\section{Impact Statement}
Our work in this paper focuses on the development of a novel benchmark, designed to evaluate the systematic visual imagination capabilities of vision and machine learning models. Our benchmark uses procedurally generated scenes and does not involve sensitive personal data or human workers. While our benchmark targets the development of more robust and effective machine learning models, we urge that models tested and improved upon using our benchmark should follow strong ethical standards. They should refrain from deployment in potentially harmful use cases, such as surveillance or spreading misinformation. Despite these broader ethical considerations, our work does not pose immediate ethical concerns.

\section{Reproducibility Statement}
Our experimental setup was implemented using PyTorch \citep{pytorch}, with each experiment requiring less than 20GB of GPU memory and concluding within a two-day timeframe. The complete benchmark and the code used to generate the benchmark as well as that used to conduct experiments is made accessible via our project page: \url{https://systematic-visual-imagination.github.io}.

\clearpage

\renewcommand{\arraystretch}{1.2}

\begin{table}[t]
\centering
\caption{{\textbf{LPIPS Performance of SVIB-dSprites.} We report the model performances on SVIB-dSprites for three levels of difficulty: Easy, Medium, and Hard, corresponding to $\alpha$ values of 0.6, 0.4, and 0.2, respectively.}}
\makebox[\textwidth]{
\begin{tabular}{L{1.3cm}L{1.8cm}@{\hskip 0.0in}C{2.2cm}@{\hskip 0.0in}@{\hskip 0.0in}C{2.2cm}@{\hskip 0.0in}@{\hskip 0.0in}C{2.2cm}@{\hskip 0.0in}@{\hskip 0.0in}C{2.2cm}@{\hskip 0.0in}@{\hskip 0.0in}C{1.5cm}@{\hskip 0.0in}}
\noalign{\smallskip}\noalign{\smallskip}
\toprule
 & Models & \textbf{Single\newline Atomic} & \textbf{Multiple Atomic} & \textbf{Single Non-Atomic} & \textbf{Multiple Non-Atomic} & \textbf{Average} \\
\midrule

\multirow{5}{*}{Easy}   & I2I-CNN   & 0.0304 & 0.0209 & 0.1524 & 0.0536 & 0.0643\\
                        & I2I-ViT   & 0.0036 & 0.0057 & 0.0050 & 0.0191 & 0.0084\\
\cdashline{2-7}[3pt/3pt]\noalign{\vskip 0.5ex}
                        & SSM-VAE   & 0.0796 & 0.1203 & 0.1499 & 0.1638 & 0.1284\\
                        & SSM-Slot  & 0.0274 & 0.0892 & 0.0228 & 0.0740 & 0.0534\\
\cdashline{2-7}[3pt/3pt]\noalign{\vskip 0.5ex} 
                        & Oracle    & 0.0006 & 0.0003 & 0.0005 & 0.0009 & 0.0006\\
\midrule
\multirow{5}{*}{Medium} & I2I-CNN   & 0.1209 & 0.0844 & 0.1703 & 0.1249 & 0.1251\\
                        & I2I-ViT   & 0.0172 & 0.0133 & 0.0535 & 0.0597 & 0.0359\\
\cdashline{2-7}[3pt/3pt]\noalign{\vskip 0.5ex} 
                        & SSM-VAE   & 0.1251 & 0.1591 & 0.1483 & 0.1735 & 0.1515\\
                        & SSM-Slot  & 0.0206 & 0.1541 & 0.0280 & 0.1241 & 0.0817\\
\cdashline{2-7}[3pt/3pt]\noalign{\vskip 0.5ex} 
                        & Oracle    & 0.0007 & 0.0004 & 0.0006 & 0.0007 & 0.0006\\
\midrule
\multirow{5}{*}{Hard}   & I2I-CNN   & 0.2514 & 0.2780 & 0.2336 & 0.3082 & 0.2678\\
                        & I2I-ViT   & 0.1700 & 0.2999 & 0.1361 & 0.3243 & 0.2326\\
\cdashline{2-7}[3pt/3pt]\noalign{\vskip 0.5ex} 
                        & SSM-VAE   & 0.2398 & 0.2748 & 0.1975 & 0.2519 & 0.2410\\
                        & SSM-Slot  & 0.1022 & 0.2759 & 0.1324 & 0.2229 & 0.1834\\
\cdashline{2-7}[3pt/3pt]\noalign{\vskip 0.5ex} 
                        & Oracle    & 0.0069 & 0.0021 & 0.0156 & 0.0010 & 0.0064\\
\bottomrule

\end{tabular}}
\label{table:SVIB-dSprites}
\end{table}

\begin{table}[t]
\centering
\caption{{\textbf{LPIPS Performance of SVIB-CLEVR.} We report the model performances on SVIB-CLEVR for three levels of difficulty: Easy, Medium, and Hard, corresponding to $\alpha$ values of 0.6, 0.4, and 0.2, respectively.}}
\makebox[\textwidth]{
\begin{tabular}{L{1.3cm}L{1.8cm}@{\hskip 0.0in}C{2.2cm}@{\hskip 0.0in}@{\hskip 0.0in}C{2.2cm}@{\hskip 0.0in}@{\hskip 0.0in}C{2.2cm}@{\hskip 0.0in}@{\hskip 0.0in}C{2.2cm}@{\hskip 0.0in}@{\hskip 0.0in}C{1.5cm}@{\hskip 0.0in}}
\noalign{\smallskip}\noalign{\smallskip}
\toprule
 & Models & \textbf{Single\newline Atomic} & \textbf{Multiple Atomic} & \textbf{Single Non-Atomic} & \textbf{Multiple Non-Atomic} & \textbf{Average} \\
\midrule

\multirow{5}{*}{Easy}   & I2I-CNN   & 0.0506 & 0.0380 & 0.0683 & 0.0802 & 0.0593\\
                        & I2I-ViT   & 0.0487 & 0.0566 & 0.0468 & 0.0811 & 0.0583\\
\cdashline{2-7}[3pt/3pt]\noalign{\vskip 0.5ex} 
                        & SSM-VAE   & 0.0602 & 0.0768 & 0.0648 & 0.1034 & 0.0763\\
                        & SSM-Slot  & 0.0288 & 0.0552 & 0.0839 & 0.0706 & 0.0596\\
\cdashline{2-7}[3pt/3pt]\noalign{\vskip 0.5ex} 
                        & Oracle    & 0.0101 & 0.0113 & 0.0110 & 0.0103 & 0.0107\\
\midrule
\multirow{5}{*}{Medium} & I2I-CNN   & 0.0967 & 0.1238 & 0.1166 & 0.1692 & 0.1265\\
                        & I2I-ViT   & 0.1313 & 0.0848 & 0.1425 & 0.1466 & 0.1263\\
\cdashline{2-7}[3pt/3pt]\noalign{\vskip 0.5ex} 
                        & SSM-VAE   & 0.0664 & 0.0962 & 0.0914 & 0.1303 & 0.0961\\
                        & SSM-Slot  & 0.0605 & 0.0471 & 0.0517 & 0.0540 & 0.0533\\
\cdashline{2-7}[3pt/3pt]\noalign{\vskip 0.5ex} 
                        & Oracle    & 0.0116 & 0.0247 & 0.0084 & 0.0118 & 0.0141\\
\midrule
\multirow{5}{*}{Hard}   & I2I-CNN   & 0.1730 & 0.2502 & 0.2080 & 0.2410 & 0.2181\\
                        & I2I-ViT   & 0.1899 & 0.2301 & 0.2005 & 0.2405 & 0.2153\\
\cdashline{2-7}[3pt/3pt]\noalign{\vskip 0.5ex} 
                        & SSM-VAE   & 0.1440 & 0.2604 & 0.1620 & 0.1871 & 0.1884\\
                        & SSM-Slot  & 0.0973 & 0.1991 & 0.0630 & 0.0960 & 0.1139\\
\cdashline{2-7}[3pt/3pt]\noalign{\vskip 0.5ex} 
                        & Oracle    & 0.0124 & 0.0131 & 0.0113 & 0.0119 & 0.0122\\
\bottomrule

\end{tabular}
}
\label{table:SVIB-CLEVR}
\end{table}

\begin{table}[t]
\centering
\caption{{\textbf{LPIPS Performance of SVIB-CLEVRTex.} We report the model performances on SVIB-CLEVRTex for three levels of difficulty: Easy, Medium, and Hard, corresponding to $\alpha$ values of 0.6, 0.4, and 0.2, respectively.}}
\makebox[\textwidth]{
\begin{tabular}{L{1.3cm}L{1.8cm}@{\hskip 0.0in}C{2.2cm}@{\hskip 0.0in}@{\hskip 0.0in}C{2.2cm}@{\hskip 0.0in}@{\hskip 0.0in}C{2.2cm}@{\hskip 0.0in}@{\hskip 0.0in}C{2.2cm}@{\hskip 0.0in}@{\hskip 0.0in}C{1.5cm}@{\hskip 0.0in}}
\noalign{\smallskip}\noalign{\smallskip}
\toprule
 & Models & \textbf{Single\newline Atomic} & \textbf{Multiple Atomic} & \textbf{Single Non-Atomic} & \textbf{Multiple Non-Atomic} & \textbf{Average} \\
\midrule

\multirow{5}{*}{Easy}   & I2I-CNN   & 0.2803 & 0.2550 & 0.3642 & 0.2757 & 0.2938\\
                        & I2I-ViT   & 0.2293 & 0.2120 & 0.3381 & 0.2322 & 0.2529\\
\cdashline{2-7}[3pt/3pt]\noalign{\vskip 0.5ex} 
                        & SSM-VAE   & 0.3504 & 0.3978 & 0.3832 & 0.3962 & 0.3819\\
                        & SSM-Slot  & 0.3044 & 0.3595 & 0.3442 & 0.3634 & 0.3429\\
\cdashline{2-7}[3pt/3pt]\noalign{\vskip 0.5ex} 
                        & Oracle    & 0.1538 & 0.1268 & 0.1514 & 0.1334 & 0.1414\\
\midrule
\multirow{5}{*}{Medium} & I2I-CNN   & 0.2751 & 0.2671 & 0.3678 & 0.2957 & 0.3014\\
                        & I2I-ViT   & 0.2277 & 0.2300 & 0.3354 & 0.2524 & 0.2614\\
\cdashline{2-7}[3pt/3pt]\noalign{\vskip 0.5ex} 
                        & SSM-VAE   & 0.3452 & 0.3917 & 0.3817 & 0.3971 & 0.3789\\
                        & SSM-Slot  & 0.2906 & 0.3449 & 0.3578 & 0.3494 & 0.3357\\
\cdashline{2-7}[3pt/3pt]\noalign{\vskip 0.5ex} 
                        & Oracle    & 0.1543 & 0.1306 & 0.1492 & 0.1434 & 0.1444\\
\midrule
\multirow{5}{*}{Hard}   & I2I-CNN   & 0.2819 & 0.3107 & 0.3651 & 0.3264 & 0.3210\\
                        & I2I-ViT   & 0.2625 & 0.3017 & 0.3145 & 0.3013 & 0.2950\\
\cdashline{2-7}[3pt/3pt]\noalign{\vskip 0.5ex} 
                        & SSM-VAE   & 0.3440 & 0.3866 & 0.3758 & 0.3916 & 0.3745\\
                        & SSM-Slot  & 0.3106 & 0.3946 & 0.3546 & 0.3762 & 0.3590\\
\cdashline{2-7}[3pt/3pt]\noalign{\vskip 0.5ex} 
                        & Oracle    & 0.1566 & 0.1221 & 0.1473 & 0.1356 & 0.1404\\
\bottomrule

\end{tabular}
}
\label{table:SVIB-CLEVRTex}
\end{table}

\renewcommand{\arraystretch}{1}
\begin{sidewaystable}[t]
\centering
\caption{\textbf{MSE Performance.} Numerical values of model performances on all 12 tasks: single atomic (S-A), single non-atomic (S-NA), multiple atomic (M-A), and multiple non-atomic (M-NA) tasks for each subset i.e., SVIB-dSprites, SVIB-CLEVR and SVIB-CLEVRTex.}
\vspace{2mm}
\resizebox{\textwidth}{!}{
\begin{tabular}{cl C{1.2cm} C{1.2cm} C{1.2cm} C{1.2cm} C{1.2cm} C{1.2cm} C{1.2cm} C{1.2cm} C{1.2cm} C{1.2cm} C{1.2cm} C{1.2cm}}
\toprule
            &           & \multicolumn{4}{c}{SVIB-dSprites}  & \multicolumn{4}{c}{SVIB-CLEVR}     & \multicolumn{4}{c}{SVIB-CLEVRTex}  \\
\cmidrule(lr){3-6}\cmidrule(lr){7-10}\cmidrule(lr){11-14}
$\alpha$    & Models    & S-A   & M-A   & S-NA  & M-NA  & S-A   & M-A   & S-NA  & M-NA  & S-A   & M-A   & S-NA  & M-NA \\
\midrule

\multirow{5}{*}{0.0}
&I2I-CNN                 & 2493.24 &   3201.78   & 2581.57     & 4273.86    & 884.86  & 1566.50 & 899.68    & 1520.91   & 647.37    & 752.06    & 988.72    & 945.42                         \\
&I2I-ViT                 & 1242.31 &   3479.05   & 1367.78     & 4220.45    & 637.14  & 1642.70 & 634.05    & 1485.16   & 591.58    & 687.23    & 1020.89   & 932.65                         \\
\cdashline{2-14}[3pt/3pt]\noalign{\vskip 0.5ex} 
&SSM-VAE                 & 2562.39 &   4300.64   & 2862.57     & 4228.19    & 975.61  & 1546.94 & 959.04    & 1330.32   & 624.35    & 1048.46   & 713.09    & 1019.86                         \\
&SSM-Slot                & 2367.21 &   2557.13   & 2588.47     & 3232.08    & 897.26  & 1582.44 & 851.21    & 1187.62   & 643.18    & 1038.00   & 721.13    & 912.87                         \\
\cdashline{2-14}[3pt/3pt]\noalign{\vskip 0.5ex}
&Oracle                  & 3798.89 &   3042.49   & 3822.39     & 2561.94    & 1366.79 & 1132.49 & 1116.75   & 851.99    & 817.70    & 983.48    & 944.21    & 936.89                         \\
\midrule

\multirow{5}{*}{0.2}
&I2I-CNN                 & 1424.42 &   2193.89   & 1643.04     & 2922.01    & 402.27  & 1050.83 & 485.51    & 800.93    & 465.28    & 655.86    & 717.17    & 680.10                         \\
&I2I-ViT                 & 843.12 &   2081.99   & 738.00     & 3015.15      & 438.48  & 974.72  & 461.85    & 808.84    & 495.31    & 614.44    & 738.57    & 666.23                         \\
\cdashline{2-14}[3pt/3pt]\noalign{\vskip 0.5ex}
&SSM-VAE                 & 1389.60 &   2027.40   & 1227.31     & 2169.14    & 312.09  & 1100.34 & 349.47    & 542.07    & 540.06    & 871.26    & 657.42    & 819.19                         \\
&SSM-Slot                & 594.88 &   1862.77   & 659.48     & 1681.50      & 194.07  & 784.53  & 126.25    & 185.84    & 411.69    & 698.89    & 518.68    & 682.04                         \\
\cdashline{2-14}[3pt/3pt]\noalign{\vskip 0.5ex}
&Oracle                  & 37.34 &   11.61   & 67.42     & 5.79             & 14.04   & 13.01   & 9.13      & 10.98     & 85.46     & 71.16     & 67.91     & 65.98                         \\
\midrule

\multirow{5}{*}{0.4}
&I2I-CNN                 & 855.21 &   600.41   & 1424.66     & 1167.54      & 208.13  & 467.12  & 243.76    & 512.15    & 384.37    & 662.02    & 522.80    & 554.77                         \\
&I2I-ViT                 & 94.72 &   70.60  & 238.69     & 514.59           & 299.26  & 327.33  & 315.14    & 417.57    & 322.02    & 562.63    & 428.36    & 446.58                            \\
\cdashline{2-14}[3pt/3pt]\noalign{\vskip 0.5ex}
&SSM-VAE                 & 632.20 &   933.52   & 850.20     & 1213.25       & 118.25  & 300.71  & 181.64    & 313.25    & 536.78    & 894.49    & 655.99    & 829.35                         \\
&SSM-Slot                & 167.65 &   710.60   & 199.35     & 673.60        & 104.18  & 99.62   & 90.80     & 99.37     & 338.61    & 632.98    & 463.35    & 554.50                         \\
\cdashline{2-14}[3pt/3pt]\noalign{\vskip 0.5ex}
&Oracle                  & 5.18 &   2.09   & 4.12     & 4.26                & 10.65   & 41.54   & 6.79      & 11.82     & 78.14     & 71.79     & 78.43     & 68.96                         \\
\midrule

\multirow{5}{*}{0.6}
&I2I-CNN                 & 164.47 &  131.91   & 1237.73     & 456.57        & 104.72  & 84.50   & 125.46    & 186.00    & 383.99    & 615.90    & 391.05    & 418.67                         \\
&I2I-ViT                 & 24.72 &   20.64   & 13.55     & 218.48           & 74.18   & 103.54  & 68.12     & 144.00    & 254.16    & 535.51    & 249.46    & 284.72                         \\
\cdashline{2-14}[3pt/3pt]\noalign{\vskip 0.5ex}
&SSM-VAE                 & 391.55 &   671.98   & 813.27     & 1180.91       & 103.26  & 193.08  & 112.05    & 228.15    & 516.31    & 840.56    & 725.84    & 805.82                         \\
&SSM-Slot                & 164.16 &   274.67   & 159.92     & 361.76        & 45.31   & 91.23   & 172.18    & 121.40    & 319.21    & 536.25    & 427.24    & 511.08                         \\
\cdashline{2-14}[3pt/3pt]\noalign{\vskip 0.5ex}
&Oracle                  & 4.54 &   1.85   & 3.44     & 4.89                & 9.86    & 11.51   & 9.15      & 9.76      & 76.99     & 74.22     & 71.38     & 64.50                         \\

\bottomrule

\end{tabular}
}

\label{table:all-reference-mse}
\end{sidewaystable}

\end{document}